\PassOptionsToPackage{table}{xcolor}
\documentclass[10pt,twocolumn,letterpaper]{article}

\usepackage{cvpr}

\definecolor{cvprblue}{rgb}{0.21,0.49,0.74}
\usepackage[pagebackref,breaklinks,colorlinks,allcolors=cvprblue]{hyperref}
\newcommand{\myparagraph}[1]{\vspace{2pt}\noindent{\bf #1}}
\usepackage{array,multirow,graphicx}
\usepackage{float}
\usepackage{amsmath}
\usepackage{mathtools}
\usepackage{bbm}
\usepackage{hhline}
\usepackage{boldline}
\usepackage{algorithm} 
\usepackage{algpseudocode}
\usepackage{graphicx}
\usepackage{amssymb}
\usepackage{colortbl}
\usepackage{booktabs}
\usepackage[accsupp]{axessibility}
\usepackage[table]{xcolor}
\definecolor{maroon}{cmyk}{0,0.87,0.68,0.32}
\usepackage{amssymb}
\usepackage{pifont}
\usepackage{cuted}
\usepackage{graphicx}
\usepackage{pifont}
\newcommand{\subplus}{{
  \resizebox{\fontcharwd\scriptfont0`0}{!}{$\scriptstyle+$}%
}}

\def\paperID{1902}
\def\confName{CVPR}
\def\confYear{2025}

\title{Spatial Transport Optimization by Repositioning Attention Map \\
for Training-Free Text-to-Image Synthesis}

\author{
Woojung Han\quad Yeonkyung Lee\quad Chanyoung Kim\quad Kwanghyun Park\quad Seong Jae Hwang\thanks{Corresponding author}\\
Yonsei University
\\{\tt\small \{dnwjddl, yeonkyung.lee, chanyoung, 
kwanghyun.park, seongjae\}@yonsei.ac.kr}
}

\begin{document}

\twocolumn[{
\maketitle
\begin{center}
    \vspace{-5pt}
    \centering
    \captionsetup{type=figure}
\includegraphics[width=\textwidth]{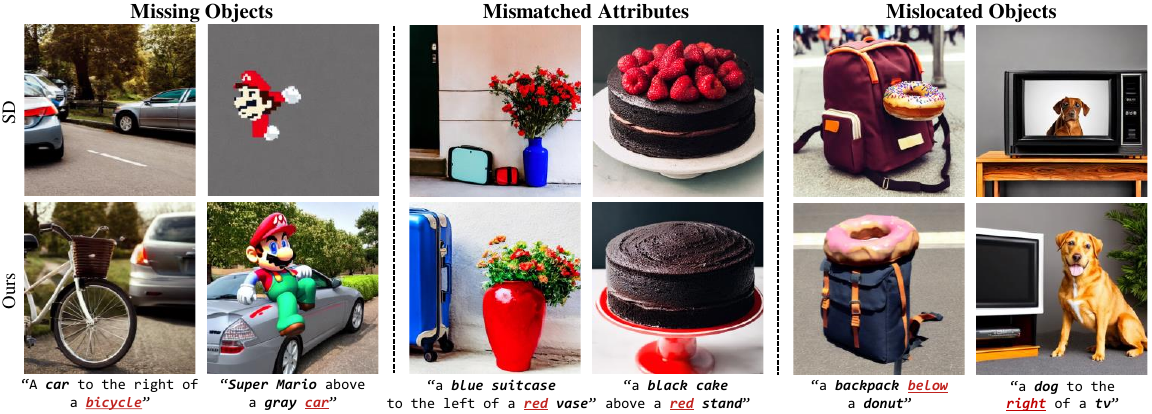}
\vspace{-20pt}
\captionof{figure}
{Three main challenges in training-free text-to-image (T2I) generation: (1) missing objects, (2) mismatched attributes, and (3) mislocated objects. While existing approaches address missing objects and mismatched attributes, effectively controlling object positioning remains an open problem. Our proposed model, \textbf{STORM}, introduces a dynamic approach to aligning relative object positions throughout the denoising process, enabling precise spatial control without additional spatial templates. \textbf{\textcolor{maroon}{\underline{Red}}} underline highlights errors made by SD.}
\label{fig:fig1}
\end{center}
}]

\begin{abstract}
Diffusion-based text-to-image (T2I) models have recently excelled in high-quality image generation, particularly in a training-free manner, enabling cost-effective adaptability and generalization across diverse tasks. However, while the existing methods have been continuously focusing on several challenges such as ``missing objects'' and ``mismatched attributes,'' another critical issue of ``mislocated objects'' remains where generated spatial positions fail to align with text prompts. \let\thefootnote\relax\footnote{
\scriptsize{Project Page:~\url{https://micv-yonsei.github.io/storm2025/}}}
Surprisingly, ensuring such seemingly basic functionality remains challenging in popular T2I models due to the inherent difficulty of imposing explicit spatial guidance via text forms. To address this, we propose \textbf{STORM} (Spatial Transport Optimization by Repositioning Attention Map), a novel training-free approach for spatially coherent T2I synthesis. 
STORM employs Spatial Transport Optimization (STO), rooted in optimal transport theory, to dynamically adjust object attention maps for precise spatial adherence, supported by a Spatial Transport (ST) Cost function that enhances spatial understanding. 
Our analysis shows that integrating spatial awareness is most effective in the early denoising stages, while later phases refine details. Extensive experiments demonstrate that STORM surpasses existing methods, effectively mitigating mislocated objects while improving missing and mismatched attributes, setting a new benchmark for spatial alignment in T2I synthesis.
\end{abstract}

\vspace{-17pt}
\section{Introduction}
\label{sec:intro}
\vspace{-7pt}
Diffusion-based text-to-image (T2I) models exhibit remarkable versatility across various applications, generating high-quality images from textual prompts~\cite{sd, sd_cdm, glide, zhang2023adding, han2024advancing}.
Notably, training-free methods offer efficiency, real-time adaptability, lower costs, and broad task generalization~\cite{attend_excite,structured_diffusion,layoutguidance}.
These training-free methods leverage their advantages while addressing three key challenges in fully capturing textual details.
In particular, as shown in Fig.~\ref{fig:fig1}, these challenges include (1) \textit{missing objects}, (2) \textit{mismatched attributes}, and (3) \textit{mislocated objects}.
These limitations may lead to images that poorly represent the intended scene or convey misleading information, reducing their reliability.

\begin{figure}[t!]
  \centering
  \includegraphics[width=\linewidth]{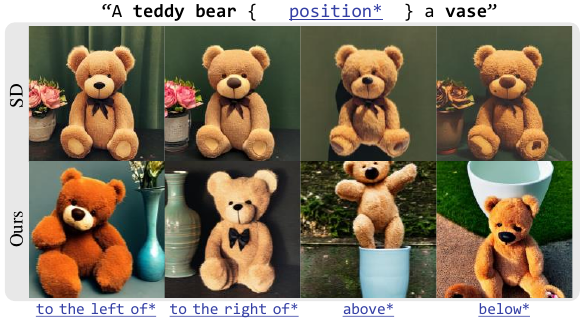}
  \vspace{-20pt}
   \caption{Comparison of Spatial Awareness.
   $\{ \ \texttt{position$^*$} \ \}$ in each prompt denotes the spatial relationship in each column (e.g., ``\textit{to the left of}'', ``\textit{to the right of}'', ``\textit{above}, and ``\textit{below}''). While Stable Diffusion (SD) shows limited spatial awareness by generating similar images regardless of spatial prompts, our model accurately reflects specified positions. (Same seed for all synthesis).
   }
  \vspace{-15pt}
   \label{fig:fig2}
\end{figure}

While there has been substantial progress in addressing issues like \textit{missing objects} and \textit{mismatched attributes} in diffusion-based image synthesis~\cite{attend_excite, divide_bind, conform, guo2024initno}, the challenge of \textit{mislocated objects} remains relatively unexplored despite its critical importance.
For example, in Fig.~\ref{fig:fig1}, the fifth column shows that the \texttt{backpack} should be below a \texttt{donut}, but Stable Diffusion (SD)~\cite{sd} generates \texttt{donut} to the left of a \texttt{backpack}.
Furthermore, SD consistently generates similar images regardless of spatial cues in the text (\textit{e.g.,} same seed with only positional changes for \texttt{teddy bear} and \texttt{vase} in Fig.~\ref{fig:fig2}), indicating limited spatial understanding.
This limitation highlights the importance of addressing the problem of mislocated objects and demonstrates the need for more advanced methods of interpreting and aligning with spatial cues.
Overcoming these challenges is key to improving T2I model reliability and supporting adaptable applications across diverse settings.

Existing solutions for \textit{mislocated objects} often rely on fixed spatial templates or predefined layouts~\cite{layoutguidance,rw_layout1,rw_layout2}, which offer some control but impose rigid constraints and require additional inputs.
For instance, these templates may fix an object’s position in the image (\textit{e.g.,} far-left corner) instead of offering flexible guidance like placing one object ``to the left'' of another.
While effective in structured fields like automated design, these approaches present limitations in more dynamic areas, such as digital art or interactive media, by restricting spatial flexibility.
Therefore, instead of relying on restrictive templates or requiring additional inputs, we aim to push the boundaries of T2I models by enabling precise alignment with textual guidance, which is essential to unlocking their full potential across applications.

\begin{figure}[t!]
  \centering
  \includegraphics[width=\linewidth]{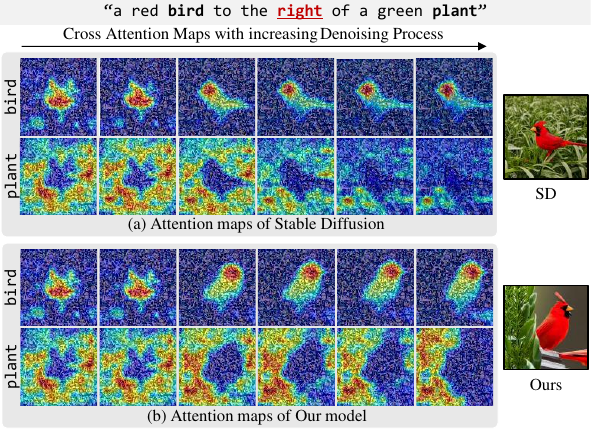}
  \vspace{-18pt}
  \caption{Comparison of Attention Map Progression During Denoising. Visualization of the attention maps for ``\textit{a red bird to the right of a green plant}'' throughout the denoising process for both Stable Diffusion (a) and our model (b). While Stable Diffusion struggles to distinctly capture the spatial relationship between the bird and the plant, our model effectively aligns the objects according to the specified spatial cue (``\textit{to the right of}''). The resulting image from our model demonstrates improved spatial accuracy compared to Stable Diffusion.}
  \vspace{-17pt}
  \label{fig:fig3}
\end{figure}

To this end, we propose \textbf{S}patial \textbf{T}ransport \textbf{O}ptimization by \textbf{R}epositioning Attention \textbf{M}ap, \textbf{STORM}, a training-free approach that adjusts relative object positions dynamically, without relying on fixed spatial templates.
While SD disregards spatial cues (see Fig.~\ref{fig:fig2}), STORM achieves various object positioning with the same model weights, ensuring spatial alignment.
To implement STORM, we introduce the Spatial Transport Optimization (STO) framework, which interprets each text token's attention map as a distribution, and leverages Optimal Transport (OT)~\cite{ot} to efficiently reposition distributions. 
At an early diffusion timestep, STO adjusts the placement of objects based on their relative positioning (see Fig.\ref{fig:fig3} for attention map changes across the denoising process).
While OT considers both distribution and distance to minimize transformation costs, we propose the Spatial Transport Cost (ST Cost) to align with our objectives and guide distribution positioning.
This function updates and optimizes the latent vector through backward guidance, following methods from prior work~\cite{attend_excite, layoutguidance}.

Our analysis suggests that integrating spatial awareness yields optimal results in the early denoising stages, whereas the later stages are better suited for focusing on fine details.
Furthermore, we found that missing objects can be resolved concurrently as mislocated objects are corrected in the early steps.
Extensive experiments show that our approach outperforms existing methods, setting a new benchmark for spatial alignment in diffusion models.

Our main contributions are as follows:
\begin{itemize}
\item{We propose STORM, a training-free method that dynamically adjusts relative object positions, achieving accurate spatial alignment without relying on fixed templates.}
\item{We introduce the Spatial Transport Optimization (STO) framework based on Optimal Transport (OT), which treats attention maps as distributions and repositions objects relative to each other across diffusion timesteps.}
\item{We develop ST Cost function that penalizes overlaps and progressively enhances spatial accuracy, demonstrating its effectiveness through extensive experiments that set a new benchmark for spatial alignment in T2I models.}
\end{itemize}
\vspace{-5pt}
\section{Related Work}
\label{sec:related_work}
\vspace{-5pt}
\paragraph{Text-to-Image Synthesis.}
Text-to-image (T2I) synthesis aims to generate visually realistic images that align with given text prompts.
Early studies focused on GANs~\cite{isola2017image, park2019semantic, karras2019style} and autoregressive models~\cite{ramesh2021zero, yu2022scaling}, but diffusion models~\cite{ho2020denoising} have taken the lead, demonstrating remarkable performance in T2I synthesis~\cite{sd,sd_cdm}.
Despite advances driven by VLMs~\cite{sd, saharia2022photorealistic, yu2022scaling}, aligning generated images with text prompts remains a challenge.
To address this, some approaches~\cite{balaji2022ediff, ramesh2022hierarchical, saharia2022photorealistic} have expanded network architectures, though these methods require retraining and are difficult to adapt to existing frameworks.
Alternatively, other research~\cite{structured_diffusion, conform, attend_excite, divide_bind} focused on training-free improvement strategies, which aligns closely with our approach.

\vspace{20pt}
\noindent\textbf{Training-free T2I Models.}
Previous studies~\cite{sd_cdm, attend_excite, structured_diffusion, divide_bind, syngen, guo2024initno, agarwal2023star} have explored various approaches to enhance how text prompts are aligned within text-to-image diffusion models.
Composable Diffusion~\cite{sd_cdm} and Structure Diffusion~\cite{structured_diffusion} aim to improve compositionality through multi-model or cross-attention methods, yet they still face issues like subject mixing and attribute binding.
To address these, Attend-and-Excite~\cite{attend_excite} dynamically adjusts attention during inference to capture all subjects, while Divide-and-Bind~\cite{divide_bind} improves attribute binding and compositional accuracy with targeted loss functions.
CONFORM~\cite{conform} uses a contrastive method to separate objects in attention maps while keeping related attributes together, and INITNO~\cite{guo2024initno} further tackles spatial misalignment by improving control over object placement to reduce overlap.
Although these studies focus on the missing object issue by reducing object overlap, they often overlook the crucial task of capturing accurate \textit{relative positioning and directional relationships} between objects.
\vspace{-18pt}
\paragraph{Training-free T2I Models for Mislocated Objects.}
The advantages of attention maps for controlling spatial positioning have led to several studies using fixed templates, such as layouts, to improve object placement~\cite{rw_layout1, rw_layout2, layoutguidance, rw_layout4, rw_layout5, nie2024compositional}.
Wu et al.~\cite{rw_layout1} use layout predictors to assign pixel regions, while ZestGuide~\cite{rw_layout2} and Chen et al.~\cite{layoutguidance} leverage cross-attention layers for alignment with segmentations or user-defined layouts.
However, these methods often require additional layout inputs and lack flexibility for relative positioning. 
As a result, there is a need for methods that enhance object placement without relying on predefined layouts.

\begin{figure*}[h]
    \centering
    \includegraphics[width=\textwidth]{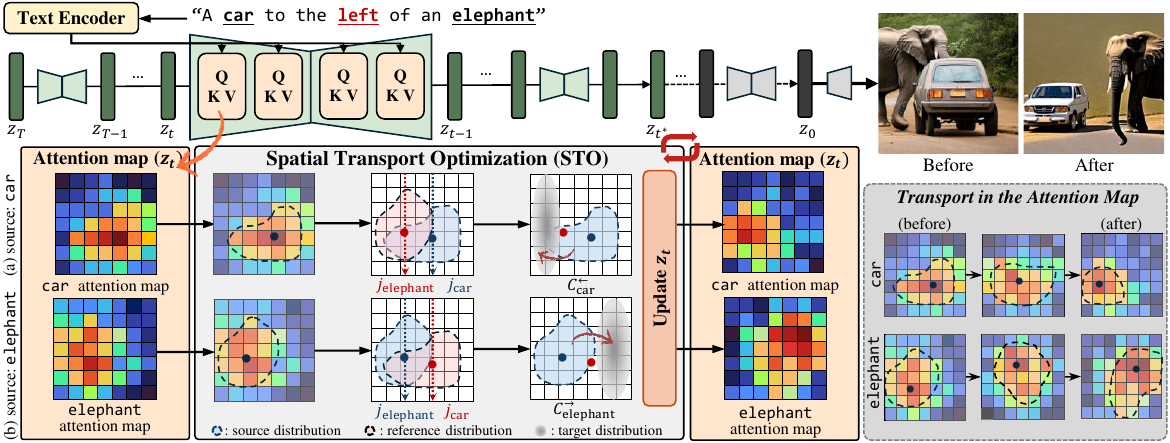}
    \vspace{-20pt}
    \caption{Overview pipeline of \textbf{STORM}.
    Our method leverages Optimal Transport in a training-free manner, allowing the model to accurately reflect relative object positions at each step without additional inputs.
    Given the prompt ``\textit{A car to the left of an elephant}'', our method dynamically adjusts the attention maps to induce the specified spatial relationship.
    The process starts with initial attention maps for the ``\textit{car}'' and ``\textit{elephant}'' at time step $z_t$. Using the centroids of these attention maps, the Spatial Transport Optimization (STO) computes the losses to correct positional relationships (\textit{e.g.,} ensuring the car is to the left of the elephant). The updated attention map is then used to refine the latent representation $z_t$, leading to a final image that adheres to the desired spatial arrangement. The comparison of attention maps (before and after STO) shows improved alignment, effectively placing the car to the left of the elephant as instructed in the prompt.}
    \vspace{-10pt}
   \label{fig:fig4}
\end{figure*}

\vspace{-7pt}
\section{Preliminaries}
\label{sec:preliminaries}
\vspace{-6pt}
\paragraph{Text Conditioned via Cross-Attention.}
Our approach uses Stable Diffusion (SD)~\cite{sd}, which generates images from text by integrating text conditioning through cross-attention.
A pre-trained CLIP encoder~\cite{radford2021learning} converts the text prompt into a conditioning vector $c$, fed into the UNet model~\cite{ronneberger2015u} via cross-attention layers at different resolutions.
At each timestep $t$, the attention map $A_t \in \mathbb{R}^{P \times P \times N}$ assigns probabilities for transferring token information to spatial positions, where $P$ is the spatial dimension of the feature map and $N$ the number of tokens in the text prompt.
In our work, we use $16 \times 16$ attention maps as they contain the most semantically meaningful information~\cite{hertz2022prompt}.
\vspace{-10pt}
\paragraph{Optimal Transport with Sinkhorn Algorithm.}
Optimal Transport (OT)~\cite{ot} provides a framework for transforming one probability distribution into another while preserving spatial relationships.
Given two discrete distributions $\mu = \{\mu_u\}_{u=1}^n$ and $\nu = \{\nu_v\}_{v=1}^n$, each with $n$ discrete locations, and a cost matrix $\mathbf{C} \in \mathbb{R}^{n \times n}$, where $\mathbf{C}_{uv}$ is the cost of moving distribution from point $u$ to $v$, OT seeks a transport plan $\mathbf{P}\in \mathbb{R}^{n \times n}$ that minimizes the total cost, $\min_{{P} \geq 0} \sum_{u=1}^n \sum_{v=1}^n \mathbf{C}_{uv} \mathbf{P}_{uv}$.
For large $n$, the Sinkhorn algorithm~\cite{sinkhorn} applies entropic regularization for a faster, scalable, and differentiable solution:
{\small
\begin{equation}
\label{eq:sinkhorn}
\vspace{-3pt}
\min_{{P} \geq 0} \quad \sum_{u=1}^n \sum_{v=1}^n \mathbf{C}_{uv} \mathbf{P}_{uv} - \varepsilon \sum_{u=1}^n \sum_{v=1}^n \mathbf{P}_{uv} (\log \mathbf{P}_{uv} - 1),   
\vspace{-3pt}
\end{equation}
}
where $\varepsilon > 0$ balances transport cost and entropy.
We further integrate spatial alignment objectives into this OT-based approach in our framework.
\vspace{-5pt}
\section{Method}
\label{sec:method}
\vspace{-6pt}
We describe the full pipeline of our model, STORM, as shown in Fig.~\ref{fig:fig4}. 
To support the understanding of our Spatial Transport Optimization (STO) framework, we first provide background information in Sec.~\ref{sec:key}. 
The full details of STO in Sec.~\ref{sec:sto}, while Sec.~\ref{sec:non-spatial} covers optimizations for non-spatial text prompts, and Sec.~\ref{sec:Optimization} details the update process.

\vspace{-3pt}
\subsection{Background of STO} 
\label{sec:key}
\vspace{-5pt}
Since objects are generated in regions of high attention within the attention map (see Fig.~\ref{fig:fig3}), our goal is to reposition the attention map based on the positional information from the text prompt, using relative objects as a reference.
To achieve this, our approach involves three types of attention maps: the attention map of the object to be moved, called the \textit{source distribution} (\textit{e.g.,} blue area in Fig.~\ref{fig:fig4}); the attention map of the relative object, which acts as a \textit{reference distribution} (\textit{e.g.,} red area in Fig.~\ref{fig:fig4}); and the \textit{target distribution} which represents the intended position for the source distribution to move (\textit{e.g.,} gray area in Fig.~\ref{fig:fig4}).
The target distribution is an arbitrary distribution defined based on the reference point, which is the centroid of the reference distribution, $(j_{\text{ref}}, i_{\text{ref}})$.
It primarily serves to guide the intended position of the source distribution using reference points without exerting significant direct influence.

\vspace{-5pt}
\subsection{Spatial Transport Optimization Framework}
\vspace{-5pt}
\label{sec:sto}
In this section, we detail our Spatial Transport Optimization (STO) framework, which dynamically repositions attention maps for accurate object alignment without modifying model weights.
This framework consists of two main components: a cost function and a transport plan.
The cost function directs the distribution shift from the source in the intended direction, allowing the attention map to adjust positionally according to spatial relationships.
Although we build on standard cost functions that measure distance between locations, our Spatial Transport Cost (ST Cost) incorporates additional objectives adapted to our specific spatial alignment goals.
The transport plan, which optimally moves the distribution between points, is implemented using the Sinkhorn algorithm.
Beginning with an initial attention map, STO iteratively updates the latent vector by applying STO-derived losses, gradually guiding the distribution into the target alignment.

\vspace{-12pt}
\subsubsection{Construction of Target Distributions}
\label{sec:target}
\vspace{-7pt}
Our approach focuses on relative positioning adjustments at each timestep, using an arbitrary target distribution (\textit{e.g.,} gray area in Fig.~\ref{fig:fig4}) to guide the source distribution based on the reference.
Depending on the text-specified direction (\textit{e.g.,} horizontal or vertical), only the relevant reference point of $(j_{\text{ref}}, i_{\text{ref}})$ is considered: for `left' or `right' positioning, only the horizontal dimension of the reference point $j_\text{ref}$ is used, while for `above' or `below' positioning, only the vertical dimension $i_\text{ref}$ is considered, allowing the flexibility in the opposite dimension.
For example, in Fig.~\ref{fig:fig4}a, with the \texttt{car} attention map as a source distribution and the \texttt{elephant} attention map as a reference distribution, the target distribution is placed to the left of a \texttt{elephant}'s distribution.
This approach similarly applies to another source.
Supplementary Materials provide details on computing the target distribution and reference point.
\vspace{-15pt}
\subsubsection{Spatial Transport Cost Function}
\vspace{-9pt}
In the STO framework, we aim to find an optimal transport plan that minimizes the cost of transferring the attention map. The cost function measures the expense of moving distribution mass by computing distances between points, where a higher value indicates less desirable locations. While our overall cost function merges the standard OT loss with an additional term (detailed in the supplementary material), here we focus on our newly introduced ST Cost, which is built on two core principles:
\vspace{-1pt}

\noindent\textbf{(I) Positional Cost:} 
We developed a cost function to guide the distribution to its intended position relative to the reference point.
In simple terms, when the source distribution is intended to be on the left side of the reference point, it has a low cost when positioned on the left and a high cost when positioned on the right.
To quantify this directional positioning, we define a set of positional $\delta$ values, representing the distances between the current point and the reference point in each of the four directions. Specifically, the $\delta$ values are defined as follows:
$\delta_{ij}^{\leftarrow} = j_\text{ref}-j,  \delta_{ij}^{\rightarrow} = j - j_\text{ref}, 
\delta_{ij}^{\uparrow} = i_\text{ref}-i, 
\delta_{ij}^{\downarrow} = i - i_\text{ref},$ with $j$ and $i$ representing the current position values, and the arrows indicate the desired and restricted direction for positioning the source attention map relative to the reference point.
When constructing the overall cost function that incorporates all directional $\delta$ values, each direction is categorized as either desired or restricted.
The desired direction ($\delta^{\text{des}}$) corresponds to the intended movement of the source distribution, while the restricted direction ($\delta^{\text{res}}$) represents the opposite or undesired direction.
For instance, if the source distribution should move to the left, $\delta_{ij}^{\leftarrow}$ is marked as the desired direction ($\delta^{\text{des}}$), while $\delta_{ij}^{\rightarrow}$ is treated as the restricted direction ($\delta^{\text{res}}$).
The combined cost function, accounting for both desired and restricted positions, can be expressed as:
\vspace{-5pt}
\begin{equation}\small
    \Delta_{ij}(\delta_{ij}^{\text{des}}, \delta_{ij}^{\text{res}}) = \frac{1}{\omega (\delta_{ij}^{\text{des}}+\epsilon)} \mathbbm{1}_\subplus[\delta_{ij}^{\text{des}}] + \omega (\delta_{ij}^{\text{res}}+\epsilon)\mathbbm{1}_\subplus[\delta_{ij}^{\text{res}}],
    \vspace{-2pt}
\end{equation}
where $\omega(\cdot) > 1$ is a progressive adaptive weight that controls the alignment importance across timesteps, and $\epsilon$ is a stabilizing factor. 
$\mathbbm{1}_\subplus$ is an indicator function that equals 1 when the input is positive and 0 otherwise, allowing selective application of computation based on positivity.
As the source distribution aligns more closely with the desired direction, the $\Delta$ value approaches zero, indicating minimal cost for correctly aligned positioning (see Example below).

\vspace{4pt}
\noindent\textbf{(II) Non-overlap Cost:}
In this process, it is crucial to prevent the source distribution from overlapping with the reference distribution during its movement.
This is one of the primary issues in addressing the missing object problem in SD, where avoiding overlap is essential for ensuring that each object is distinctly represented.
To achieve this, we simply embed the reference distribution directly into the cost function, guiding objects to occupy separate locations.
Specifically, with $A_{ij}$ representing the attention weight of the reference distribution, we incorporated this distribution into the cost function, assigning high costs to regions occupied by the reference distribution and low costs elsewhere.

Combining this principle with our core ideas, our cost function can be defined as follows:
{\small
\begin{equation}
\vspace{-1pt}
C_{ij} = {A}_{ij} \Delta_{ij}(\delta_{ij}^{\text{des}}, \delta_{ij}^{\text{res}}).
\vspace{-8pt}
\end{equation}
}

\paragraph{Example.}
In Fig.~\ref{fig:fig4}, the car should be left of the elephant, and the elephant to the right of the car.
The cost functions for each of them are represented by $C_{ij}^\leftarrow={A}_{ij} \Delta_{ij}(\delta_{ij}^{\leftarrow}, \delta_{ij}^{\rightarrow})$, $C_{ij}^\rightarrow={A}_{ij} \Delta_{ij}(\delta_{ij}^{\rightarrow}, \delta_{ij}^{\leftarrow})$, respectively.
Taking the car as an example, it needs to be positioned to the left of the elephant relative to the elephant's centroid point.
In more detail, if the current point is positioned to the left of the centroid, the $\delta^{\text{des}}_{ij}$ is activated; if positioned to the right of the centroid, the $\delta^{\text{res}}_{ij}$ is activated.
Additionally, as the car moves to the left of the elephant, it should not overlap the elephant it passes along the way.
The elephant's attention map is used in the cost function, guiding the car to avoid overlapping with the elephant's area.
The resulting $C_{ij}$ is utilized as the cost function for STO.
\begingroup
\renewcommand{\arraystretch}{0.9}
\begin{table*}[t!] 
\centering
\caption{Performance comparison between different models on VISOR (\%) and Object Accuracy (OA) (\%) metrics, based on Stable Diffusion 1.4 and Stable Diffusion 2.1~\cite{sd}. For entries marked with $\dagger$, official values were unavailable; thus, we directly calculated these values. \textbf{Bold} values indicate the best scores, while \underline{underlined} values indicate the second-best scores among the fair comparison.
\textit{Notes on VISOR.} VISOR cond: spatial accuracy using the results in which both objects are sufficiently generated; VISOR uncond: spatial accuracy using all results, including results with failed object generations; 
VISOR 1/2/3/4: ratios of one, two, three, or four spatially accurate results out of four seed results, respectively. (e.g., VISOR$_1$: \% of results with exactly one spatially correct result out of four generated results)}
\vspace{-10pt}
\label{table:main_results}
\resizebox{\textwidth}{!}{
\begin{tabular}{lccccccccccc}
\toprule 
\multirow{2}{*}{{Model}} & \multirow{2}{*}{{Venue}} & \multirow{2}{*}{{Extra Training}} & \multirow{2}{*}{{Extra Input}} & \multirow{2}{*}{{OA (\%)}} & \multicolumn{6}{c}{{VISOR (\%)}} \\
\cmidrule(lr){6-11} 
& & & & & {uncond} & {cond} & {1} & {2} & {3} & {4} \\
\midrule 
\rowcolor{gray!20} 
\textit{Stable Diffusion 1.4 Based} &  & &   &   &    &   &   &   &   &  \\ 
SD 1.4 \cite{sd} & & - & \ding{55} & 29.86 & 18.81 & 62.98 & 46.60 & 20.11 & 6.89 & 1.63 \\ 
SD 1.4 + CDM \cite{sd_cdm} & \textcolor{gray}{\scriptsize ECCV22} & \ding{51} & \ding{55} & 23.27 & 14.99 & 64.41 & 39.44 & 14.56 & 4.84 & 1.12 \\
GLIDE \cite{glide} & \textcolor{gray}{\scriptsize ICML22} & \ding{51} & \ding{55} & 3.36 & 1.98 & 59.06 & 6.72 & 1.02 & 0.17 & 0.03 \\
GLIDE + CDM \cite{sd_cdm} &\textcolor{gray}{\scriptsize ECCV22}  & \ding{51} & \ding{55} & 10.17 & 6.43 & 63.21 & 20.07 & 4.69 & 0.83 & 0.11 \\
Control-GPT \cite{zhang2023controllable} &\textcolor{gray}{\scriptsize arXiv23} & \ding{51} & \ding{51} & {48.33} & 44.17 & 65.97 & {69.80} & {51.20} & {35.67} & {20.48} \\
Layout Guidance \cite{layoutguidance} & \textcolor{gray}{\scriptsize WACV24} & \ding{55} & \ding{51} & 40.01 & {38.80} & {95.95} & - & - & - & - \\
REVISION \cite{chatterjee2025revision} & \textcolor{gray}{\scriptsize ECCV24} & \ding{55} & \ding{51} & 53.96 & 52.71 & 97.69 & 77.79 & 61.02 & 44.90 & 27.15 \\ \hline
Structure Diffusion \cite{structured_diffusion} & \textcolor{gray}{\scriptsize ICLR23} & \ding{55} & \ding{55} & 28.65 & 17.87 & 62.36 & 44.70 & 18.73 & 6.57 & 1.46 \\
Attend-and-Excite~\cite{attend_excite} & \textcolor{gray}{\scriptsize SIGGRAPH23} & \ding{55} & \ding{55} & 42.07 & 25.75 & 61.21 & 49.29 & 19.33 & 4.56 & 0.08 \\ 
Divide-and-Bind$^\dagger$~\cite{divide_bind} & \textcolor{gray}{\scriptsize BMVC24} & \ding{55} & \ding{55} & 46.03 & 31.62 & \underline{68.70} & 64.72 & 37.82 & 18.64 & 5.30 \\ 
INITNO$^\dagger$ \cite{guo2024initno} & \textcolor{gray}{\scriptsize CVPR24} & \ding{55} & \ding{55} & 60.40 & 35.18 & 58.24 & 71.20 & 42.71 & 20.09 & 6.72 \\ 
CONFORM$^\dagger$ \cite{conform} & \textcolor{gray}{\scriptsize CVPR24} & \ding{55} & \ding{55} & \underline{60.73} & \underline{38.48} & 62.33 & \underline{73.01} & \underline{45.82} & \underline{25.57} & \underline{9.52} \\ 
\rowcolor{blue!5}
{Ours (SD 1.4)} & & \textbf{\ding{55}} & \textbf{\ding{55}} & \textbf{61.01} & \textbf{57.58} & \textbf{94.39} & \textbf{85.93} & \textbf{69.71} & \textbf{49.01} & \textbf{25.70} \\  
\midrule  
\midrule
\rowcolor{gray!20} 
\textit{Stable Diffusion 2.1 Based} & &  &   &   &    &   &   &   &   &  \\ 
SD 2.1~\cite{sd} &  & - & \ding{55} & 47.83 & {30.25} & 63.24 & 64.42 & 35.74 & 16.13 & 4.70 \\ 
SPRIGHT \cite{chatterjee2025getting} & \textcolor{gray}{\scriptsize ECCV24}  & \ding{51} & \ding{55} & 60.68 & 42.23 & 71.24 & 71.78 & 51.88 & 33.09 & 16.15 \\ 
REVISION \cite{chatterjee2025revision} & \textcolor{gray}{\scriptsize ECCV24}  & \ding{55} & \ding{51} & 48.26 & 47.11 & 97.61 & 76.07 & 55.75 & 37.10 & 19.53 \\ \hline 
\rowcolor{blue!5}
{Ours (SD 2.1)} & & \textbf{\ding{55}} & \textbf{\ding{55}} & \textbf{62.55} & \textbf{59.35} & \textbf{94.88} & \textbf{88.34} & \textbf{71.75} & \textbf{52.03} & \textbf{25.42} \\
\bottomrule 
\end{tabular}
}
\vspace{-10pt}
\end{table*}
\endgroup
\vspace{-7pt}
\subsubsection{Sinkhorn Algorithm-based Transport Plan}
\vspace{-7pt}
After combining the standard OT approach with ST components (see Supplementary Material for the complete formulation), we compute the transport plan using the Sinkhorn algorithm~\cite{sinkhorn}, which applies entropic regularization for computational efficiency and stability, especially for high-dimensional attention maps. 
The regularized OT problem is formulated as in Eq.~\eqref{eq:sinkhorn} using target distribution calculated in Sec.~\ref{sec:target}.
The Sinkhorn algorithm iteratively updates transport plan $\textbf{P}$ to meet the marginal constraints of source and target distributions, applying the cost function bidirectionally for both objects.

\vspace{-6pt}
\subsection{Optimization for Objects and Attributes}
\label{sec:non-spatial}
\vspace{-6pt}
Our method also extends to cases where the text lacks spatial information.
(i) Object-level Optimization: We handle cases where the text prompt specifies objects without explicit spatial guidance by randomly placing the target distribution of each object.
This flexible positioning encourages diverse object placement when spatial relationships are not provided.
Therefore, the cost function is simplified to focus on solely on preventing overlap, expressed as
${C_{ij}} = {A}_{ij} \otimes \textbf{1}_N$, where $A$ is the reference attention map.
(ii) Attribute Matching Optimization: The attention map for an object and its corresponding attribute should be identical, reflecting the same attention focus.
Therefore, we used the standard OT cost function, calculating Euclidean distance between corresponding positions across the object, and attribute attention maps to quantify differences.

\vspace{-5pt}
\subsection{Optimization and Update Process}
\label{sec:Optimization}
\vspace{-5pt}
We update the latent vector using STO-based losses to guide the spatial alignment of the attention maps in a differentiable manner. 
At each timestep, we compute the OT loss $\mathcal{L}$ and update the latent vector $z_t$ as $ z'_{t} \leftarrow z_t - \alpha_t \cdot \nabla{z_t} \mathcal{L} $ where $\alpha_t$ is the step size.
After the update, a forward pass computes $z_{t-1}$ for the next denoising step.

\vspace{2pt}
\noindent\textbf{Remark.}
As shown in Fig.\ref{fig:ablation}, addressing the mislocated object problem is most effective in early timesteps, which, according to related work\cite{park2024explaining}, establish the overall image structure while later timesteps refine finer details.
Based on this, STORM establishes object positioning and spatial configurations early in the denoising process, minimizing the risk of misplacements and overlaps.
Furthermore, as shown in the bottom row of Fig.~\ref{fig:ablation}, solving the mislocated object problem before addressing missing objects can naturally resolve the latter, highlighting the importance of prioritizing mislocated objects as a critical step in the synthesis process.
\vspace{-3pt}
\vspace{-8pt}
\section{Experiments}
\label{sec:experiments}
\vspace{-4pt}
\noindent\textbf{Implementation Details.}
Following methods established in prior research~\cite{attend_excite, divide_bind, conform}, we conducted a comprehensive evaluation of our model. 
We used the official Stable Diffusion models~\cite{sd} v1.4 and v2.1, configuring the denoising timestep to 50, and the guidance scale to 7.5 in line with common practices in the field. 
To smooth the cross-attention map, a Gaussian filter with a kernel size of 3 and a standard deviation of 0.5 was applied.
Optimization was performed iteratively with forward updates at specified timesteps ($t \in {5,10,15,20}$), stopping at $t=25$ to enhance quality. 
All experiments use NVIDIA RTX A6000 $\text{Ada}^{*}$.
\let\thefootnote\relax\footnote{
\scriptsize{$^{*}$Advanced Database System Infrastructure(NFEC-2024-11-300458) 
}}
\subsection{Evaluation Results}
\vspace{-5pt}
We evaluated our method against existing training-free T2I models using both qualitative and quantitative measures, focusing on two key aspects.
(i) Spatial Accuracy: To measure alignment with spatial configurations specified in prompts, we used the VISOR~\cite{visor} and the T2I-CompBench~\cite{t2icompbench}, both provide a comprehensive assessment of spatial relationships captured by the model.
(ii) Object and Attribute Consistency: We assessed object and attribute representation using T2I-CompBench in combination with CLIP-space distance~\cite{radford2021learning} and TIFA score~\cite{hu2023tifa}, following established methods~\cite{attend_excite, divide_bind, conform}. We provide detailed descriptions of each evaluation metric and datasets in the Supplementary Materials for additional clarity.

\myparagraph{Quantitative Evaluation: VISOR.}
To evaluate spatial accuracy, we use the VISOR~\cite{visor} to measure the ability of a model to position objects based on spatial cues (\textit{e.g.,} above) in text prompts. 
VISOR benchmark contains over 25K prompts describing spatial relationships between objects.
Notably, as seen in Table~\ref{table:main_results}, the increase in $\text{VISOR}_4$ demonstrates the model's reliability in generating spatially accurate images across prompts and seeds.

\begingroup
\renewcommand{\arraystretch}{1.0}
\begin{table}[t!]
\centering
\caption{Comparison of methods on T2I-CompBench, with attribute binding and spatial relationships calculated using models based on Stable Diffusion 1.4 and 2.1.}
\vspace{-10pt}
\label{table:t2icompbench}
\resizebox{\columnwidth}{!}{ 
\begin{tabular}{lcccc}
\toprule
Method & \begin{tabular}[c]{@{}c@{}}Attribute\\Binding\\(Color ↑)\end{tabular} & \begin{tabular}[c]{@{}c@{}}Attribute\\Binding\\(Shape ↑)\end{tabular} & \begin{tabular}[c]{@{}c@{}}Attribute\\Binding\\(Texture ↑)\end{tabular} & \begin{tabular}[c]{@{}c@{}}Object\\Relationship\\(Spatial ↑)\end{tabular} \\
\midrule
\rowcolor{gray!20}
\textit{Stable Diffusion 1.4 Based} & & & & \\
Stable Diffusion-v1.4~\cite{sd}    & 0.3765 & 0.3576 & 0.4156 & 0.1246  \\
\rowcolor{blue!5}
\textbf{Ours (SD 1.4)} & \textbf{0.6458} & \textbf{0.5983} & \textbf{0.7539} & \textbf{0.1613} \\
\rowcolor{gray!20}
\textit{Stable Diffusion 2.1 Based} & & & & \\
Stable Diffusion-v2.1~\cite{sd}    & 0.5065 & 0.4221  & 0.4922 & 0.1342  \\
Composable Diffusion~\cite{sd_cdm}  & 0.4063 & 0.3299 & 0.3645 & 0.0800  \\
Structured Diffusion~\cite{structured_diffusion}  & 0.4990 & 0.4218 & 0.4900 & 0.1386 \\
Attend-and-Excite~\cite{attend_excite}   & 0.6400 & 0.4517 & 0.5963 & 0.1455  \\ 
\rowcolor{blue!5}
\textbf{Ours (SD 2.1)} & \textbf{0.6777} & \textbf{0.6226} & \textbf{0.7884} & \textbf{0.1981}  \\
\bottomrule
\end{tabular}
}
\vspace{-16pt}
\end{table}
\endgroup

\vspace{3pt}
\myparagraph{Quantitative Evaluation: T2I-CompBench.}
The T2I-CompBench~\cite{t2icompbench} evaluates object presence, spatial relationships, and attribute alignment from text prompts.
We used the attribute and spatial configuration datasets provided by T2I-CompBench, averaging results over more than 10 randomly selected seeds for each dataset.
Attribute Binding is evaluated with the BLIP-VQA metric~\cite{li2022blip}, which poses separate questions for each attribute-object pair in the prompt.
UniDet-based metric~\cite{unidet} is used for spatial relationships. 
In Table~\ref{table:t2icompbench}, our score shows that the generated images position the specified objects correctly and accurately reflect the given attributes (e.g., color, shape, texture).

\begingroup
\renewcommand{\arraystretch}{0.9}
\begin{table}[h!]
\centering
\vspace{-3pt}
\caption{User study evaluating model performance on object synthesis, attribute matching, spatial correctness, and overall fidelity.}
\vspace{-8pt}
\label{table:userstudy}
\resizebox{\columnwidth}{!}{ 
\begin{tabular}{lcccc}
\toprule
Method & \begin{tabular}[c]{@{}c@{}}Object\\Accuracy (\%)\end{tabular} & \begin{tabular}[c]{@{}c@{}}Attribute\\Matching (\%)\end{tabular} & \begin{tabular}[c]{@{}c@{}}Spatial\\Correctness (\%)\end{tabular} & \begin{tabular}[c]{@{}c@{}}Overall\\Fidelity (\%)\end{tabular} \\
\midrule
Stable Diffusion~\cite{sd}    & 14.68 & 14.31 & 11.81 & 15.02  \\
Attend-and-Excite~\cite{attend_excite}    & 16.34 & 16.50  & 13.20 & 16.04  \\
Divide-and-Bind~\cite{divide_bind}  & 16.39 & 15.69 & 13.01 & 16.52  \\
INITNO~\cite{guo2024initno}   & 15.98 & 16.03 & 13.15 & 15.63  \\ 
CONFORM~\cite{conform}  & 15.64 & 16.79 & 12.91 & 14.14 \\
\rowcolor{blue!5}
\textbf{Ours} & \textbf{20.97} & \textbf{20.68} & \textbf{35.92} & \textbf{22.65}  \\
\bottomrule
\end{tabular}
}
\vspace{-13pt}
\end{table}
\endgroup

\begin{figure*}[h]
    \centering
    \includegraphics[width=\textwidth]{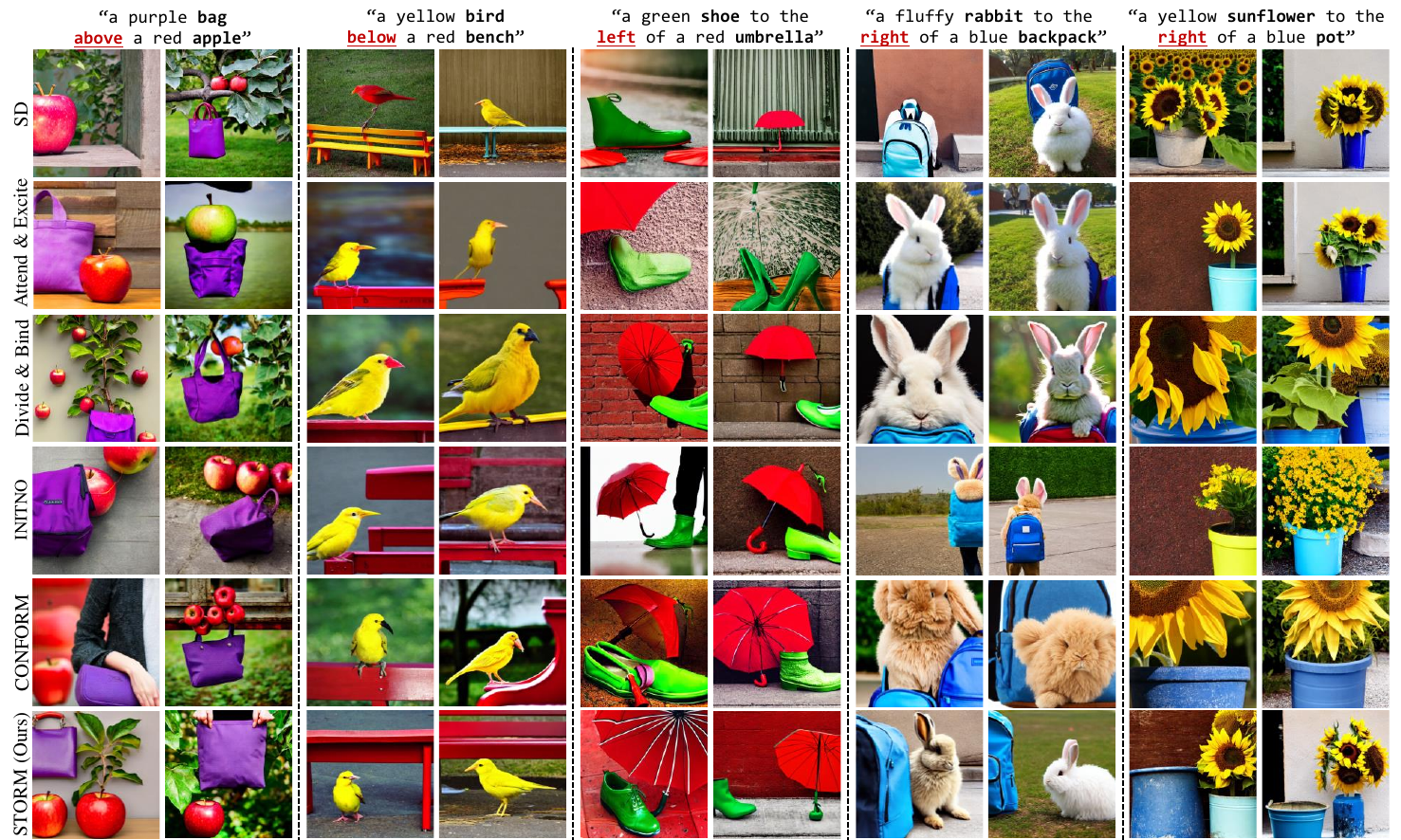}
    \vspace{-20pt}
    \caption{Qualitative comparison across the custom prompt, which involves attribute and positional information in text, evaluating previous state-of-the-art training-free T2I methods, Attend\&Excite~\cite{attend_excite}, Divide\&Bind~\cite{divide_bind}, INITNO~\cite{guo2024initno}, CONFORM~\cite{conform}, and ours.}
    \label{fig:fig6}
    \vspace{-10pt}
\end{figure*}
\myparagraph{Quantitative Evaluation: CLIP Score and TIFA Score.}
Following previous work~\cite{attend_excite, conform, guo2024initno}, we evaluate our method using CLIP Image-Text and Text-Text Similarity, generating 64 images per prompt and averaging CLIP cosine similarity scores across methods.
Fig.~\ref{fig:clip} reports two CLIP Image-Text Similarity metrics: full prompt similarity (cosine similarity between the entire prompt and generated image) and minimum object similarity (minimum similarity between the generated image and each subject prompt) to see the accuracy of each object. 
We generate captions for the images with BLIP~\cite{li2022blip} and compute cosine similarity with the original prompt. 
Our method achieves comparable performance to other leading methods.
However, as the reliability of the CLIP score is limited by its ``bag-of-words'' approach~\cite{yuksekgonul2022and} (as noted by \cite{attend_excite}), we also use the TIFA metric~\cite{hu2023tifa} for improved evaluation of complex prompts.
TIFA~\cite{hu2023tifa} evaluates image faithfulness to the input prompt by generating question-answer pairs using the LLaMA 2~\cite{touvron2023llama} model.
Our method performs well compared to other state‐of‐the‐art models.
\begin{figure}[t!]
  \centering
  \includegraphics[width=0.95\linewidth]{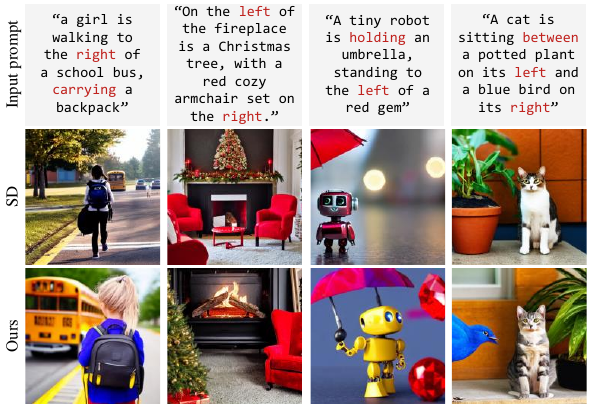}
  \vspace{-8pt}
   \caption{Generated images with complex positioning details in the prompt (Additional results in Supplementary Materials).}
   \label{fig:complex_prompt}
  \vspace{-14pt}
\end{figure}

\myparagraph{Quantitative Evaluation: User Studies.}
We create 10 custom text prompts with detailed objects, attributes, and spatial information, then generate images using random seeds.
With 30 participants, our user study evaluates (1) object accuracy, (2) attribute matching, (3) spatial correctness, and (4) overall fidelity.
In Table~\ref{table:userstudy}, our model shows surpassing results across the existing models of all categories.
\begin{figure*}[t!]
  \centering
  \includegraphics[width=\linewidth]{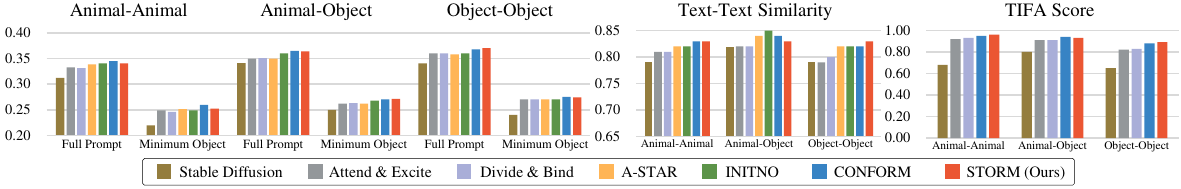}
  \vspace{-20pt}
   \caption{Comparison of various models on CLIP Image-Text, CLIP Text-Text, and TIFA scores across prompts involving Animal-Animal, Animal-Object, and Object-Object pairs. (The seed is randomly selected.)}
   \label{fig:clip}
   \vspace{-13pt}
\end{figure*}
\begin{figure}[t!]
  \centering
  \includegraphics[width=\linewidth]{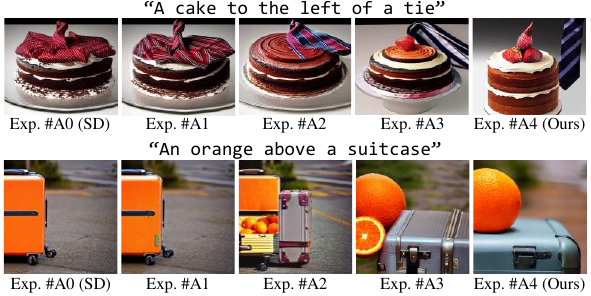}
  \vspace{-20pt}
   \caption{Comparison of results when applying STO at different timesteps. Experiments are organized as follows: no STO, STO applied from timesteps 19-24, 13-24, 7-24, and 1-24 (Ours). As seen in the images, earlier STO application improves object positioning and reduces overlap, resulting in more accurately positioned objects. (See Table~\ref{table:timestep} for quantitative evaluation.)}
   \label{fig:ablation}
  \vspace{-15pt}
\end{figure}

\myparagraph{Qualitative Analysis.}
Fig.~\ref{fig:fig6} provides a qualitative comparison with recent state-of-the-art models~\cite{conform,guo2024initno}. 
To compare attribute matching and object localization accuracy, we create custom text prompts to evaluate the qualitative performance of our model. 
Existing methods often misalign spatial details, leading to errors such as a \texttt{rabbit} overlapping a \texttt{backpack} instead of being positioned to its right.
In contrast, our method, which is focused on precise positioning, demonstrates a strong understanding of spatial information and generates objects accurately.
Our method also can generate images that accurately reflect complex spatial information, even when prompts contain intricate positioning details (refer Fig.~\ref{fig:complex_prompt}). 

\vspace{-4pt}
\subsection{Ablation Studies}
\vspace{-5pt}

\noindent\textbf{Effects of Optimizing through Timestep.}
\label{sec:timestep}
In Table~\ref{table:timestep}, we compare STO applied at different timestep ranges (Exp.\#A1: 19–24, Exp.\#A2: 13–24, Exp.\#A3: 7–24, Exp.\#A4: 1–24).
These experiments show that setting spatial configurations in the early stages is essential not only for positional accuracy but also for achieving higher OA. 
This improvement increases as spatial relationships are established earlier in the process.
Notably, all configurations with STO outperform the baseline (Exp.\#A0, without STO), demonstrating a substantial improvement in both object positioning and accuracy.
In Fig.~\ref{fig:ablation}, tie placement is optimal when spatial cues are applied early, while delayed guidance leads to misplacements or suboptimal synthesis.

\vspace{2pt}
\myparagraph{Effects of Cost Function.} 
Table~\ref{table:costfunction} presents an ablation study on methods for computing movement costs between a source and target distribution.
Exp.\#B0 uses Manhattan distance (only horizontal and vertical movement), while Exp.\#B1 considers the mass distribution of the attention map using Euclidean distance.
Exp.\#B2 combines Exp.\#B0 and Exp.\#B1.
Exp.\#B3 excludes the overlap prevention component from ST Cost to examine its impact.
Our results (Exp.\#B4) show that ST Cost outperforms standard OT functions, with overlap prevention proving critical for maintaining OA, despite minimal impact on VISOR.
\begingroup
\renewcommand{\arraystretch}{0.9}
\begin{table}[t!]
\centering
\caption{Ablation study on the impact of applying STO at different timesteps. Exp.\#A0 shows baseline results from SD without STO. From Exp.\#A1 to Exp.\#A4, STO is progressively applied across increasing timestep ranges (19-24, 13-24, 7-24, and 1-24).}
\vspace{-8pt}
\label{table:timestep}
\resizebox{\columnwidth}{!}{
\begin{tabular}{c ccccccc}
\toprule
\multirow{2}{*}{\# Exp.} & \multirow{2}{*}{OA (\%)} & \multicolumn{6}{c}{VISOR} \\ \cmidrule(lr){3-8}
   &  & uncond & cond  & 1 & 2  & 3  & 4  \\ \midrule
A0  & 29.86  & 18.81  & 62.98 & 46.60 & 20.11 & 6.89 & 1.63   \\ 
A1  & 46.62  & 37.48  & 80.38 & 74.11 & 45.73 & 22.30 & 7.82   \\ 
A2  & 55.65  & 47.90  & 86.07 & 81.63 & 60.30 & 35.89 & 13.81   \\
A3   & 59.85  & 53.62  & 89.60 & 83.61 & 67.53 & 44.12 & 19.31   \\ 
\rowcolor{blue!5}
A4 (Ours)  & \textbf{61.01}  &  \textbf{57.58} & \textbf{94.39} & \textbf{85.93} & \textbf{69.71} & \textbf{49.01} & \textbf{25.70}    \\
\bottomrule
\end{tabular}
}
\vspace{-5pt}
\end{table}
\endgroup

\begingroup
\renewcommand{\arraystretch}{0.9}
\begin{table}[t!]
\centering
\caption{Ablation study on computing cost function, highlighting the impact of ST Cost components on OA and VISOR performance. (B0: Manhattan distance; B1: Euclidean distance; B2: B0 + B1; B3: ST cost without overlap prevention.)}
\vspace{-8pt}
\label{table:costfunction}
\resizebox{\columnwidth}{!}{
\begin{tabular}{c ccccccc}
\toprule
\multirow{2}{*}{\# Exp.} & \multirow{2}{*}{OA (\%)} & \multicolumn{6}{c}{VISOR} \\ \cmidrule(lr){3-8}
         &       & uncond   & cond  & 1      & 2     & 3     & 4   \\ \midrule
B0  & 27.32  & 17.91  & 65.51 & 42.84 & 19.25 & 7.60   & 2.00  \\
B1    & 47.48  & 30.35  & 63.93 & 65.20  & 37.71 & 14.03 & 4.54   \\ 
B2     & 46.60 & 29.20  & 62.93 & 63.92 & 33.91 & 14.43 & 4.60    \\
B3     & 31.38 & 27.78  & 88.53 & 61.30 & 32.61 & 13.44 & 3.82    \\
\rowcolor{blue!5}
B4 (Ours)  & \textbf{61.01} & \textbf{57.58} & \textbf{94.39} & \textbf{85.93} & \textbf{69.71} & \textbf{49.01} & \textbf{25.70}  \\
\bottomrule
\end{tabular}
}
\vspace{-10pt}
\end{table}
\endgroup

\vspace{-3pt}
\section{Conclusion}
\vspace{-3pt}
\label{sec:conclusion}
In this work, we introduced STORM, a framework that dynamically addresses spatial misalignment issues in training-free T2I synthesis. 
By leveraging the STO framework, which combines OT and an ST Cost function, STORM not only resolves mislocated objects but also tackles missing objects and mismatched attributes.
Extensive experiments show that STORM significantly improves spatial alignment and object accuracy, surpassing existing state-of-the-art methods.
\let\thefootnote\relax\footnote{\scriptsize{{\bf Acknowledgement.} 
This work was supported in part by IITP RS-2024-00457882 (AI Research Hub Project), IITP 2020-0-01361, NRF RS-2024-00345806,  NRF RS-2023-00219019, and RS-2024-00403860 (Korea Basic Science Institute, National Research Facilities and Equipment Center).
}}
This advancement establishes STORM as a new benchmark for spatial control in T2I generation, pushing the boundaries of adaptability and efficiency in training-free approaches.
\clearpage

{
    \small
    \bibliographystyle{ieeenat_fullname}
    \bibliography{main}

\begin{thebibliography}{46}
\providecommand{\natexlab}[1]{#1}
\providecommand{\url}[1]{\texttt{#1}}
\expandafter\ifx\csname urlstyle\endcsname\relax
  \providecommand{\doi}[1]{doi: #1}\else
  \providecommand{\doi}{doi: \begingroup \urlstyle{rm}\Url}\fi

\bibitem[Agarwal et~al.(2023)Agarwal, Karanam, Joseph, Saxena, Goswami, and Srinivasan]{agarwal2023star}
Aishwarya Agarwal, Srikrishna Karanam, KJ Joseph, Apoorv Saxena, Koustava Goswami, and Balaji~Vasan Srinivasan.
\newblock A-star: Test-time attention segregation and retention for text-to-image synthesis.
\newblock In \emph{ICCV}, 2023.

\bibitem[Balaji et~al.(2022)Balaji, Nah, Huang, Vahdat, Song, Zhang, Kreis, Aittala, Aila, Laine, et~al.]{balaji2022ediff}
Yogesh Balaji, Seungjun Nah, Xun Huang, Arash Vahdat, Jiaming Song, Qinsheng Zhang, Karsten Kreis, Miika Aittala, Timo Aila, Samuli Laine, et~al.
\newblock ediff-i: Text-to-image diffusion models with an ensemble of expert denoisers.
\newblock \emph{arXiv preprint arXiv:2211.01324}, 2022.

\bibitem[Brown(2020)]{brown2020language}
Tom~B Brown.
\newblock Language models are few-shot learners.
\newblock \emph{arXiv preprint arXiv:2005.14165}, 2020.

\bibitem[Chatterjee et~al.(2025{\natexlab{a}})Chatterjee, Luo, Gokhale, Yang, and Baral]{chatterjee2025revision}
Agneet Chatterjee, Yiran Luo, Tejas Gokhale, Yezhou Yang, and Chitta Baral.
\newblock Revision: Rendering tools enable spatial fidelity in vision-language models.
\newblock In \emph{ECCV}, 2025{\natexlab{a}}.

\bibitem[Chatterjee et~al.(2025{\natexlab{b}})Chatterjee, Stan, Aflalo, Paul, Ghosh, Gokhale, Schmidt, Hajishirzi, Lal, Baral, et~al.]{chatterjee2025getting}
Agneet Chatterjee, Gabriela Ben~Melech Stan, Estelle Aflalo, Sayak Paul, Dhruba Ghosh, Tejas Gokhale, Ludwig Schmidt, Hannaneh Hajishirzi, Vasudev Lal, Chitta Baral, et~al.
\newblock Getting it right: Improving spatial consistency in text-to-image models.
\newblock In \emph{ECCV}, 2025{\natexlab{b}}.

\bibitem[Chefer et~al.(2023)Chefer, Alaluf, Vinker, Wolf, and Cohen-Or]{attend_excite}
Hila Chefer, Yuval Alaluf, Yael Vinker, Lior Wolf, and Daniel Cohen-Or.
\newblock Attend-and-excite: Attention-based semantic guidance for text-to-image diffusion models.
\newblock \emph{SIGGRAPH}, 2023.

\bibitem[Chen et~al.(2024)Chen, Laina, and Vedaldi]{layoutguidance}
Minghao Chen, Iro Laina, and Andrea Vedaldi.
\newblock Training-free layout control with cross-attention guidance.
\newblock In \emph{WACV}, 2024.

\bibitem[Couairon et~al.(2023)Couairon, Careil, Cord, Lathuiliere, and Verbeek]{rw_layout2}
Guillaume Couairon, Marlene Careil, Matthieu Cord, St{\'e}phane Lathuiliere, and Jakob Verbeek.
\newblock Zero-shot spatial layout conditioning for text-to-image diffusion models.
\newblock In \emph{ICCV}, 2023.

\bibitem[Cuturi(2013)]{sinkhorn}
Marco Cuturi.
\newblock Sinkhorn distances: Lightspeed computation of optimal transport.
\newblock \emph{NeurIPS}, 2013.

\bibitem[Feng et~al.(2022)Feng, He, Fu, Jampani, Akula, Narayana, Basu, Wang, and Wang]{structured_diffusion}
Weixi Feng, Xuehai He, Tsu-Jui Fu, Varun Jampani, Arjun Akula, Pradyumna Narayana, Sugato Basu, Xin~Eric Wang, and William~Yang Wang.
\newblock Training-free structured diffusion guidance for compositional text-to-image synthesis.
\newblock \emph{arXiv preprint arXiv:2212.05032}, 2022.

\bibitem[Gokhale et~al.(2022)Gokhale, Palangi, Nushi, Vineet, Horvitz, Kamar, Baral, and Yang]{visor}
Tejas Gokhale, Hamid Palangi, Besmira Nushi, Vibhav Vineet, Eric Horvitz, Ece Kamar, Chitta Baral, and Yezhou Yang.
\newblock Benchmarking spatial relationships in text-to-image generation.
\newblock \emph{arXiv preprint arXiv:2212.10015}, 2022.

\bibitem[Gong et~al.(2024)Gong, Huang, Feng, Zhang, Li, and Liu]{rw_layout4}
Biao Gong, Siteng Huang, Yutong Feng, Shiwei Zhang, Yuyuan Li, and Yu Liu.
\newblock Check locate rectify: A training-free layout calibration system for text-to-image generation.
\newblock In \emph{CVPR}, 2024.

\bibitem[Guo et~al.(2024)Guo, Liu, Cui, Li, Yang, and Huang]{guo2024initno}
Xiefan Guo, Jinlin Liu, Miaomiao Cui, Jiankai Li, Hongyu Yang, and Di Huang.
\newblock Initno: Boosting text-to-image diffusion models via initial noise optimization.
\newblock In \emph{CVPR}, 2024.

\bibitem[Han et~al.(2024)Han, Kim, Ju, Shim, and Hwang]{han2024advancing}
Woojung Han, Chanyoung Kim, Dayun Ju, Yumin Shim, and Seong~Jae Hwang.
\newblock Advancing text-driven chest x-ray generation with policy-based reinforcement learning.
\newblock \emph{MICCAI}, 2024.

\bibitem[Hertz et~al.(2022)Hertz, Mokady, Tenenbaum, Aberman, Pritch, and Cohen-Or]{hertz2022prompt}
Amir Hertz, Ron Mokady, Jay Tenenbaum, Kfir Aberman, Yael Pritch, and Daniel Cohen-Or.
\newblock Prompt-to-prompt image editing with cross attention control.
\newblock \emph{arXiv preprint arXiv:2208.01626}, 2022.

\bibitem[Ho et~al.(2020)Ho, Jain, and Abbeel]{ho2020denoising}
Jonathan Ho, Ajay Jain, and Pieter Abbeel.
\newblock Denoising diffusion probabilistic models.
\newblock \emph{NeurIPS}, 2020.

\bibitem[Hu et~al.(2024)Hu, Wang, Fang, Fu, Cheng, and Yu]{hu2024ella}
Xiwei Hu, Rui Wang, Yixiao Fang, Bin Fu, Pei Cheng, and Gang Yu.
\newblock Ella: Equip diffusion models with llm for enhanced semantic alignment.
\newblock \emph{arXiv preprint arXiv:2403.05135}, 2024.

\bibitem[Hu et~al.(2023)Hu, Liu, Kasai, Wang, Ostendorf, Krishna, and Smith]{hu2023tifa}
Yushi Hu, Benlin Liu, Jungo Kasai, Yizhong Wang, Mari Ostendorf, Ranjay Krishna, and Noah~A Smith.
\newblock Tifa: Accurate and interpretable text-to-image faithfulness evaluation with question answering.
\newblock In \emph{ICCV}, 2023.

\bibitem[Huang et~al.(2023)Huang, Sun, Xie, Li, and Liu]{t2icompbench}
Kaiyi Huang, Kaiyue Sun, Enze Xie, Zhenguo Li, and Xihui Liu.
\newblock T2i-compbench: A comprehensive benchmark for open-world compositional text-to-image generation.
\newblock \emph{NeurIPS}, 2023.

\bibitem[Isola et~al.(2017)Isola, Zhu, Zhou, and Efros]{isola2017image}
Phillip Isola, Jun-Yan Zhu, Tinghui Zhou, and Alexei~A Efros.
\newblock Image-to-image translation with conditional adversarial networks.
\newblock In \emph{CVPR}, 2017.

\bibitem[Karras et~al.(2019)Karras, Laine, and Aila]{karras2019style}
Tero Karras, Samuli Laine, and Timo Aila.
\newblock A style-based generator architecture for generative adversarial networks.
\newblock In \emph{CVPR}, 2019.

\bibitem[Li et~al.(2022)Li, Li, Xiong, and Hoi]{li2022blip}
Junnan Li, Dongxu Li, Caiming Xiong, and Steven Hoi.
\newblock Blip: Bootstrapping language-image pre-training for unified vision-language understanding and generation.
\newblock In \emph{ICML}, 2022.

\bibitem[Li et~al.(2024)Li, Jain, and Shi]{divide_bind}
Jiachen Li, Jitesh Jain, and Humphrey Shi.
\newblock Matting anything.
\newblock In \emph{CVPR}, 2024.

\bibitem[Liu et~al.(2022)Liu, Li, Du, Torralba, and Tenenbaum]{sd_cdm}
Nan Liu, Shuang Li, Yilun Du, Antonio Torralba, and Joshua~B Tenenbaum.
\newblock Compositional visual generation with composable diffusion models.
\newblock In \emph{ECCV}, 2022.

\bibitem[Meral et~al.(2024)Meral, Simsar, Tombari, and Yanardag]{conform}
Tuna Han~Salih Meral, Enis Simsar, Federico Tombari, and Pinar Yanardag.
\newblock Conform: Contrast is all you need for high-fidelity text-to-image diffusion models.
\newblock In \emph{CVPR}, 2024.

\bibitem[Minderer et~al.(2022)Minderer, Gritsenko, Stone, Neumann, Weissenborn, Dosovitskiy, Mahendran, Arnab, Dehghani, Shen, et~al.]{minderer2022simple}
Matthias Minderer, Alexey Gritsenko, Austin Stone, Maxim Neumann, Dirk Weissenborn, Alexey Dosovitskiy, Aravindh Mahendran, Anurag Arnab, Mostafa Dehghani, Zhuoran Shen, et~al.
\newblock Simple open-vocabulary object detection.
\newblock In \emph{ECCV}, 2022.

\bibitem[Nichol et~al.(2021)Nichol, Dhariwal, Ramesh, Shyam, Mishkin, McGrew, Sutskever, and Chen]{glide}
Alex Nichol, Prafulla Dhariwal, Aditya Ramesh, Pranav Shyam, Pamela Mishkin, Bob McGrew, Ilya Sutskever, and Mark Chen.
\newblock Glide: Towards photorealistic image generation and editing with text-guided diffusion models.
\newblock \emph{arXiv preprint arXiv:2112.10741}, 2021.

\bibitem[Nie et~al.(2024)Nie, Liu, Mardani, Liu, Eckart, and Vahdat]{nie2024compositional}
Weili Nie, Sifei Liu, Morteza Mardani, Chao Liu, Benjamin Eckart, and Arash Vahdat.
\newblock Compositional text-to-image generation with dense blob representations.
\newblock 2024.

\bibitem[Park et~al.(2024)Park, Ju, and Lee]{park2024explaining}
Ji-Hoon Park, Yeong-Joon Ju, and Seong-Whan Lee.
\newblock Explaining generative diffusion models via visual analysis for interpretable decision-making process.
\newblock \emph{Expert Systems with Applications}, 2024.

\bibitem[Park et~al.(2019)Park, Liu, Wang, and Zhu]{park2019semantic}
Taesung Park, Ming-Yu Liu, Ting-Chun Wang, and Jun-Yan Zhu.
\newblock Semantic image synthesis with spatially-adaptive normalization.
\newblock In \emph{CVPR}, 2019.

\bibitem[Radford et~al.(2021)Radford, Kim, Hallacy, Ramesh, Goh, Agarwal, Sastry, Askell, Mishkin, Clark, et~al.]{radford2021learning}
Alec Radford, Jong~Wook Kim, Chris Hallacy, Aditya Ramesh, Gabriel Goh, Sandhini Agarwal, Girish Sastry, Amanda Askell, Pamela Mishkin, Jack Clark, et~al.
\newblock Learning transferable visual models from natural language supervision.
\newblock In \emph{ICML}, 2021.

\bibitem[Ramesh et~al.(2021)Ramesh, Pavlov, Goh, Gray, Voss, Radford, Chen, and Sutskever]{ramesh2021zero}
Aditya Ramesh, Mikhail Pavlov, Gabriel Goh, Scott Gray, Chelsea Voss, Alec Radford, Mark Chen, and Ilya Sutskever.
\newblock Zero-shot text-to-image generation.
\newblock In \emph{ICML}, 2021.

\bibitem[Ramesh et~al.(2022)Ramesh, Dhariwal, Nichol, Chu, and Chen]{ramesh2022hierarchical}
Aditya Ramesh, Prafulla Dhariwal, Alex Nichol, Casey Chu, and Mark Chen.
\newblock Hierarchical text-conditional image generation with clip latents.
\newblock \emph{arXiv preprint arXiv:2204.06125}, 2022.

\bibitem[Rassin et~al.(2024)Rassin, Hirsch, Glickman, Ravfogel, Goldberg, and Chechik]{syngen}
Royi Rassin, Eran Hirsch, Daniel Glickman, Shauli Ravfogel, Yoav Goldberg, and Gal Chechik.
\newblock Linguistic binding in diffusion models: Enhancing attribute correspondence through attention map alignment.
\newblock \emph{NeurIPS}, 2024.

\bibitem[Rombach et~al.(2022)Rombach, Blattmann, Lorenz, Esser, and Ommer]{sd}
Robin Rombach, Andreas Blattmann, Dominik Lorenz, Patrick Esser, and Bj{\"o}rn Ommer.
\newblock High-resolution image synthesis with latent diffusion models.
\newblock In \emph{CVPR}, 2022.

\bibitem[Ronneberger et~al.(2015)Ronneberger, Fischer, and Brox]{ronneberger2015u}
Olaf Ronneberger, Philipp Fischer, and Thomas Brox.
\newblock U-net: Convolutional networks for biomedical image segmentation.
\newblock In \emph{MICCAI}, 2015.

\bibitem[Saharia et~al.(2022)Saharia, Chan, Saxena, Li, Whang, Denton, Ghasemipour, Gontijo~Lopes, Karagol~Ayan, Salimans, et~al.]{saharia2022photorealistic}
Chitwan Saharia, William Chan, Saurabh Saxena, Lala Li, Jay Whang, Emily~L Denton, Kamyar Ghasemipour, Raphael Gontijo~Lopes, Burcu Karagol~Ayan, Tim Salimans, et~al.
\newblock Photorealistic text-to-image diffusion models with deep language understanding.
\newblock \emph{NeurIPS}, 2022.

\bibitem[Sun et~al.(2024)Sun, Li, Lin, and Zhang]{rw_layout5}
Wenqiang Sun, Teng Li, Zehong Lin, and Jun Zhang.
\newblock Spatial-aware latent initialization for controllable image generation.
\newblock \emph{arXiv preprint arXiv:2401.16157}, 2024.

\bibitem[Touvron et~al.(2023)Touvron, Martin, Stone, Albert, Almahairi, Babaei, Bashlykov, Batra, Bhargava, Bhosale, et~al.]{touvron2023llama}
Hugo Touvron, Louis Martin, Kevin Stone, Peter Albert, Amjad Almahairi, Yasmine Babaei, Nikolay Bashlykov, Soumya Batra, Prajjwal Bhargava, Shruti Bhosale, et~al.
\newblock Llama 2: Open foundation and fine-tuned chat models.
\newblock \emph{arXiv preprint arXiv:2307.09288}, 2023.

\bibitem[Villani et~al.(2009)]{ot}
C{\'e}dric Villani et~al.
\newblock Optimal transport: old and new.
\newblock \emph{Springer}, 338, 2009.

\bibitem[Wu et~al.(2023)Wu, Liu, Zhao, Bui, Lin, Zhang, and Chang]{rw_layout1}
Qiucheng Wu, Yujian Liu, Handong Zhao, Trung Bui, Zhe Lin, Yang Zhang, and Shiyu Chang.
\newblock Harnessing the spatial-temporal attention of diffusion models for high-fidelity text-to-image synthesis.
\newblock In \emph{ICCV}, 2023.

\bibitem[Yu et~al.(2022)Yu, Xu, Koh, Luong, Baid, Wang, Vasudevan, Ku, Yang, Ayan, et~al.]{yu2022scaling}
Jiahui Yu, Yuanzhong Xu, Jing~Yu Koh, Thang Luong, Gunjan Baid, Zirui Wang, Vijay Vasudevan, Alexander Ku, Yinfei Yang, Burcu~Karagol Ayan, et~al.
\newblock Scaling autoregressive models for content-rich text-to-image generation.
\newblock \emph{arXiv preprint arXiv:2206.10789}, 2022.

\bibitem[Yuksekgonul et~al.(2022)Yuksekgonul, Bianchi, Kalluri, Jurafsky, and Zou]{yuksekgonul2022and}
Mert Yuksekgonul, Federico Bianchi, Pratyusha Kalluri, Dan Jurafsky, and James Zou.
\newblock When and why vision-language models behave like bags-of-words, and what to do about it?
\newblock \emph{arXiv preprint arXiv:2210.01936}, 2022.

\bibitem[Zhang et~al.(2023{\natexlab{a}})Zhang, Rao, and Agrawala]{zhang2023adding}
Lvmin Zhang, Anyi Rao, and Maneesh Agrawala.
\newblock Adding conditional control to text-to-image diffusion models.
\newblock In \emph{ICCV}, 2023{\natexlab{a}}.

\bibitem[Zhang et~al.(2023{\natexlab{b}})Zhang, Zhang, Vineet, Joshi, and Wang]{zhang2023controllable}
Tianjun Zhang, Yi Zhang, Vibhav Vineet, Neel Joshi, and Xin Wang.
\newblock Controllable text-to-image generation with gpt-4.
\newblock \emph{arXiv preprint arXiv:2305.18583}, 2023{\natexlab{b}}.

\bibitem[Zhou et~al.(2022)Zhou, Koltun, and Kr{\"a}henb{\"u}hl]{unidet}
Xingyi Zhou, Vladlen Koltun, and Philipp Kr{\"a}henb{\"u}hl.
\newblock Simple multi-dataset detection.
\newblock In \emph{CVPR}, 2022.

\end{thebibliography}
}

\clearpage

\section*{A. Additional Material: Presentation Video}
We have described our results in an easily accessible manner on our project page, where a brief \textbf{presentation video} is also available. The link to the \textbf{project page} is as follows: \url{https://micv-yonsei.github.io/storm2025/}.

\section*{B. Implementation Details}
We discuss the hyperparameter settings and selection of object and attribute tokens (Section B.1 and Section B.2).

\subsection*{B.1. Hyperparameter Details}
In this section, we detail the hyperparameter settings used in our implementation, ensuring alignment with previous models for a fair comparison.
We adopt a scale factor of 20, and a scale range of (1.0, 0.5), consistent with prior work to maintain uniformity in updating the denoised latent $z_t$ across denoising steps.
Similarly, the Gaussian smoothing parameters, such as the standard deviation ($\sigma$) of 0.5 and the kernel size of 3, are set identical to those in other models.
Optimization is applied only for the first 25 timesteps, as in prior work, to prevent quality degradation in the generated image. This ensures that modifications primarily enhance spatial awareness during the critical early denoising stages while preserving overall image fidelity in later stages.
For timesteps $t \in {5,10,15,20}$, additional iterations are performed during optimization if the specific target values of 0.05, 0.01, 0.005, and 0.001 are not achieved.
The optimization process ensures that the model converges toward these precise thresholds, with a maximum of 30 iterations allowed for each timestep.

\subsection*{B.2. Selection of Object and Attribute Tokens}
We utilize a part-of-speech (POS) tagger to extract nouns (object tokens) and adjectives (attribute tokens) from the given prompt.
Additionally, users have the flexibility to manually specify tokens of interest, a method consistent with approaches employed in previous studies~\cite{guo2024initno, attend_excite, divide_bind, conform}, allowing for further customization and refinement based on specific requirements.
These tokens are then analyzed through attention maps to ensure the model focuses more effectively on the identified tokens.
For tokens conveying positional information (\textit{e.g.,} "on the left," "next to," "above"), the model leverages the extracted spatial context from the text prompt to guide its operations. The explicitly stated positional information dynamically adjusts the attention maps of both object and attribute tokens, ensuring that spatial relationships in the prompt are accurately reflected in the generated output.

\section*{C. Method Details}
In this section, we discuss details of reference point and target distribution (Section C.1 and Section C.2, respectively), details of ST Cost (Section C.3), Sinkhorn algorithm-based Transport Plan (Section C.4), and algorithm of our method (Section C.5).

\subsection*{C.1. Reference Centroid Positioning}
The reference point is defined as the centroid of the attention map for the relative object, serving as an anchor for spatial adjustments.
This centroid guides the placement of the target distribution, ensuring that the source distribution aligns accurately with the desired spatial relationship.
The centroid coordinates along the horizontal and vertical dimensions are computed as follows:
\begin{equation}\small 
\begin{aligned} 
j_{\text{A}}=\frac{\sum_{j=0}^{n-1} j \cdot \left( \sum_{i=0}^{n-1} {A}_{ij} \right)}{\sum_{i=0}^{n-1} \sum_{j=0}^{n-1} {A}_{ij}}, \ i_{\text{A}}=\frac{\sum_{i=0}^{n-1} i \cdot \left( \sum_{j=0}^{n-1} {A}_{ij} \right)}{\sum_{i=0}^{n-1} \sum_{j=0}^{n-1} {A}_{ij}}. 
\end{aligned} 
\end{equation}
Here, ${A}_{ij}$ represents the attention value at position $(i, j)$ in the attention map ${A}$.
The computed values, $j_{A}$ and $i_{A}$, correspond to the centroid positions along the horizontal and vertical axes, respectively, obtained as weighted averages over each dimension.

\subsection*{C.2. Target Distribution}
The target distribution is an arbitrary distribution representing the desired position of the source distribution relative to the reference point.
To model this, we adopt a circular Gaussian distribution, providing a probabilistic representation of the object's spatial presence.
This formulation allows the distribution to adapt dynamically based on the specified spatial constraints at each timestep.
The reference point determines the centroid of the Gaussian distribution (see Section C.1 for details on reference point computation).
To compute this centroid, we consider the relative spatial relationship (left, right, above, below) with respect to the reference point ($j_\text{ref}, i_\text{ref})$.
The centroid is defined as follows, where $N$ represents the size of one dimension of the image:
$c^{\leftarrow}=(\frac{0+j_\text{ref}}{2}, \frac{N}{2}), c^{\rightarrow}=(\frac{N+j_\text{ref}}{2}, \frac{N}{2}),
c^{\uparrow}=(\frac{N}{2}, \frac{0+i_\text{ref}}{2}),
c^{\downarrow}=(\frac{N}{2}, \frac{N+i_\text{ref}}{2}).$
Each arrow corresponds to a specific spatial direction.
After computing the centroid of the target Gaussian distribution, the final target distribution is defined as:
\begin{equation}
    D(x,y) = \text{exp}\Big(-\frac{(x-c_x)^2 + (y-c_y)^2}{2\sigma^2}\Big),
\end{equation}
where $c_x$ and $c_y$ denote the centroid coordinates along the horizontal (x) and vertical (y) dimensions, respectively.

\subsection*{C.3. Details of ST Cost}
\paragraph{Details of $\omega$.}
In Equation 2 of the main paper, $\omega$ represents the progressive adaptive weights in the Spatial Transport (ST) cost function.
It controls the trade-off between aligning the source distribution in the desired direction and penalizing movement in restricted directions.
$\omega$ is defined as:
\begin{equation}
    \label{eq:omega}
    \omega(t) = 1 + (\omega_{max} - 1) (1-e^{-kt}),
\end{equation} 
where $t$ represents the timestep and $\omega_{\text{max}}$ is the maximum weight value, and $k$ is a constant controlling the rate of increase.
In our experimental setting, we set $\omega_{\text{max}}$ to 100.
A higher $\omega$ emphasizes precise alignment in the desired direction, potentially sacrificing flexibility, whereas a lower $\omega$ may allow more flexibility but reduce positional accuracy.
It gradually increases in later steps to enforce precise positioning and ensure smooth movement that maintains image quality.
Starting with a lower $\omega$ for flexibility, it gradually increases in later steps to enforce precise positioning and ensure smooth movement that maintains image quality.
Further details on the selection of $\omega$ values are provided in the ablation studies (Section E.1.1).

\paragraph{Details of Cost.}
Since $C_{ij}$ (refer to Equation 3 in the main paper) is evaluated only along a specific dimension, we extend it to the other axis by integrating $\mathbf{1}_N$ along that dimension.
This extension is formulated as $\textbf{C}^{\text{st}} = C_{\text{flat}} \otimes \mathbf{1}_N$, where $C_{\text{flat}}$ represents the flattened version of $C_{ij}$ (refer Equation 3 in the main paper) and $N$ is the number of patches ($n \times n$).
To ensure an optimal transport plan that accounts for positional relationships, we also incorporate the standard OT cost matrix, following traditional OT formulations to quantify the spatial cost of transporting mass between distributions.
To construct this matrix, we compute the $p$-norm distance between all possible pairs of points in the source and target distributions.
First, the source and target distributions are represented as 2D grids of dimensions $H$ (height) and $W$ (width).
Each grid is then converted into a list of patch coordinates, $(i,j)$ for the source and $(k,l)$ for the target, capturing all possible spatial locations.
For each pair of coordinates $(i,j)$ and $(k,l)$, we compute the $p$-norm distance defined as:
\begin{equation}
    d_p((i,j),(k,l)) = {(|i-k|^{p} + |j-l|^p)}^{1/p}
\end{equation}
The cost matrix $\textbf{C}^{\text{dist}}$ is constructed by taking the $p$-th power of these distances, resulting in :
\begin{equation}
    \textbf{C}^{\text{dist}}_{uv} =  (|i_u - k_v|^p + |j_u- l_v|^p),
\end{equation}
where $\textbf{C}^{\text{dist}}_{uv}$ is the cost of transporting mass from the $u$-th coordinates in the source to the $v$-th coordinate in the target.
Finally, the overall cost matrix is defined as:
\begin{equation} \mathbf{C} = \lambda \mathbf{C}^{\text{dist}} + (1-\lambda) \mathbf{C}^{\text{st}}, \end{equation}
where $\lambda = 0.01$ is chosen to minimize the influence of uncertainty in the target distribution.

\noindent\textbf{Optimization for Objects.}
We simplify by removing position-based terms, focusing solely on ensuring that objects do not overlap. This is expressed as $\mathbf{C}^{\text{st}} = {A}_{\text{flat}} \otimes \textbf{1}_N$, while $\mathbf{C}^{\text{dist}}$ operates in the same manner as described above.

\subsection*{C.4. Sinkhorn Algorithm-based Transport Plan}
\label{sec:sinkhorn}
Once the cost function is established, the next step is to compute a transport plan $\mathbf{P}$ that minimizes this cost.
To solve this OT problem efficiently, we employ the Sinkhorn algorithm, an iterative approach that introduces an entropic regularization term to the standard OT objectives.
The regularization term ensures that $\mathbf{P}$ becomes more evenly distributed and computationally stable, particularly for high-dimensional attention maps.
The cost matrix $\mathbf{C}$ obtained from the customized cost function (ST Cost), encodes the spatial alignment objectives between the source and target distributions.
For instance, $\mathbf{C}_{uv}$ represents the alignment cost between source position $u$ and target position $v$.
The regularized OT problem is formulated as:
\begin{equation}
\min_{\mathbf{P} \geq 0} \quad \sum_{u=1}^N \sum_{v=1}^N \mathbf{C}_{uv} \mathbf{P}_{uv} - \lambda \sum_{u=1}^N \sum_{v=1}^N \mathbf{P}_{uv} (\log \mathbf{P}_{uv} - 1), 
\end{equation}
where $\lambda$ controls the strength of the entropic regularization.
The transport plan $\mathbf{P}$ is initialized as a uniform matrix and iteratively refined to satisfy the marginal constraints defined by the source ($A$) and target ($B$) distributions.
To satisfy the row and column marginals, $A$ and $B$, the Sinkhorn algorithm alternately updates the scaling vectors $u$ and $v$ as follows:
\begin{equation}
    \textbf{u} \xleftarrow{}\frac{A}{\textbf{Pv}},   
    \textbf{v} \xleftarrow{}\frac{B}{\textbf{P}^\top\textbf{u}}.
\end{equation}
At each step, these updates ensure that the rows and columns of $\mathbf{P}$ sum to the respective marginal distributions.
The algorithm iterates until the constraints are satisfied within a predefined tolerance.
To ensure spatial consistency, the cost function is applied bi-directionally.
If the position of $A$ is adjusted using $B$ as a reference, the reverse operation is also performed, adjusting $B$ using $A$.

\subsection*{C.5. Algorithm}
Algorithm~\ref{alg:alg} provides an overview of the denoising process using STORM, which includes the update and optimization process.

\begin{algorithm}[h!]
\caption{A Denoising Step using STORM}
\label{alg:alg}
\begin{flushleft}
\textbf{Input:} \\
- A text prompt $\mathcal{P}$\\
- Attention map keys $\mathcal{K} = \{\text{source}, \text{reference}\}$\\
- Current timestep $t$\\
- Iterations for refinement $\{t_1, \dots, t_k\}$\\
- Thresholds $\{T_1, \dots, T_k\}$\\
- Trained Stable Diffusion model $SD$\\

\textbf{Output:} \\
- A noised latent $z_{t-1}$ for the next timestep
\end{flushleft}
\begin{algorithmic}[1]
\State $\_, A_t \gets \text{SD}(z_t, \mathcal{P}, t)$ \Comment{\small{Obtain attention map $A_t$ from SD.}}
\State $A_t \gets \text{Softmax}(A_t - \langle \text{sot} \rangle)$ \Comment{\small{Apply softmax to exclude special tokens.}}

\State $\mathcal{A} \gets \{\}$ \Comment{\small{Initialize attention map dictionary.}}
\State $\mathcal{C} \gets \{\}$ \Comment{\small{Initialize centroid dictionary.}}

\For{$k \in \mathcal{K}$} \Comment{\small{Process all specified attention keys.}}
    \State $\mathcal{A}[k] \gets A_t[:, :, k]$ \Comment{\small{Extract attention map for key $k$.}}
    \State $\mathcal{C}[k] \gets \text{ComputeCentroid}(\mathcal{A}[k])$ \Comment{\small{Compute centroid for key $k$.}}
\EndFor

\State $A_t^{\text{source}} \gets \mathcal{A}[\text{source}]$
\State $c^{\text{source}} \gets \mathcal{C}[\text{source}]$

\State $A_t^{\text{ref}} \gets \mathcal{A}[\text{reference}]$
\State $c^{\text{ref}} \gets \mathcal{C}[\text{reference}]$

\State $D_{\text{target}} \gets \text{Gaussian}(c_\text{ref})$
\Comment{\small{Target Distribution based on Gaussian or Spatial Prior aligned with $c^{\text{ref}}$.}}

\State $\mathcal{C} \gets \text{ST Cost}(A_t^{\text{source}}, c^{\text{ref}}, D_{\text{target}})$ \Comment{\small{Compute cost matrix using centroids.}}
\State $\mathcal{T} \gets \text{Sinkhorn}(A_t^{\text{source}}, D_{\text{target}}, \mathcal{C})$ \Comment{\small{Compute transport plan.}}

\State $\mathcal{L} \gets \sum \mathcal{T} \cdot \mathcal{C}$ \Comment{\small{Calculate st loss.}}

\State $z_t' \gets z_t - \alpha_t \cdot \nabla_{z_t} \mathcal{L}$ \Comment{\small{Update latent $z_t$ using gradient.}}

\If{$t \in \{t_1, \dots, t_k\}$} \Comment{\small{Check if iterative refinement is needed.}}
    \If{$\mathcal{L} > 1 - T_t$} \Comment{\small{Compare loss against threshold $T_t$.}}
        \State $z_t \gets z_t'$
        \State \textbf{Go to} Step 1
    \EndIf
\EndIf

\State $z_{t-1}, \_ \gets \text{SD}(z_t', \mathcal{P}, t)$ \Comment{\small{Obtain updated latent $z_{t-1}$.}}
\State \textbf{Return} $z_{t-1}$
\end{algorithmic}
\label{alg:storm-denoising}
\end{algorithm}

\section*{D. Evaluation Metrics and Datasets}
This section provides a comprehensive overview of the three primary metrics~\cite{t2icompbench, visor,hu2023tifa}, along with the details of the user studies presented in the main paper.
Additionally, it includes descriptions of the datasets used for calculating each metric.

\subsection*{D.1. VISOR metric}
The VISOR (Verifying Spatial Object Relationships) evaluates the spatial reasoning capabilities of T2I models by assessing how accurately they generate images that reflect the spatial relationships described in text prompts.
The key components of VISOR are defined as follows:

\paragraph{Object Accuracy (OA)}
Object Accuracy (OA) measures whether both objects specified in the text prompt are present in the generated image.
It is computed as $OA(x, A, B) = \mathbbm{1}_{h(x)}(\exists A \cap \exists B)$, 1 if both $A$ and $B$ are detected in image $x$, otherwise 0.
OA measures object presence using OWL-ViT~\cite{minderer2022simple}, a pre-trained open-vocabulary object detector.

\paragraph{$\text{VISOR}_{\text{uncond}}$}
The $\text{VISOR}_{\text{uncond}}$ evaluates spatial correctness by determining whether the generated spatial relationship aligns with the ground truth relationship specified in the text prompt.
It assesses both the presence of the objects in the generated image and whether their spatial arrangement accurately reflects the prompt description.
\begin{equation} \small
\text{VISOR}_\text{uncond}(x, A, B, R) =
\begin{cases} 
1, & \text{if } (R_\text{gen} = R) \land (\exists A \cap \exists B), \\
0, & \text{otherwise},
\end{cases}    
\end{equation}
where $R_\text{gen}$ indicates the spatial relationship detected between objects in the generated image, and $R$ denotes the ground truth relationship specified in the text prompt. 
The term $\exists A \cap \exists B$ indicates that both objects $A$ and $B$ are detected in the image.
This metric provides a holistic evaluation of the model’s ability to both generate objects and accurately position them according to the specified spatial relationships. 
Unlike metrics that strictly require the detection of both objects, $\text{VISOR}_{\text{uncond}}$ captures a more comprehensive view of the model's real-world performance.
For example, given the prompt ``A cat to the left of a dog'', if the generated image contains both a cat and a dog with a cat positioned correctly to the left of a dog, then: $\text{VISOR}_\text{uncond} = 1$.
Otherwise, $\text{VISOR}_{\text{uncond}}$ = 0.

\paragraph{$\text{VISOR}_\text{cond}$}
This metric evaluates spatial correctness only when both objects are correctly generated in the image.
The spatial relationship is determined using centroid-based rules (\textit{e.g.,} $x_A < x_B$ implies $A$ is to the left of $B$).
For example, given the prompt ``A cat to the left of a dog,'' only generated images that contain both a cat and a dog are considered for evaluation. 
If the cat is correctly positioned to the left of the dog, then $\text{VISOR}_\text{cond} = 1$. Otherwise, 0.

\paragraph{$\text{VISOR}_{n}$}
$\text{VISOR}_{n}$ measures the probability of generating a least $n$ spatially correct images for a given text prompt when multiple images are generated.
If reflects a model's practical utility for users who select from multiple outputs, capturing its consistency in producing spatially accurate generations.

\paragraph{SR2D Dataset}
The SR2D (Spatial Relationships in 2D) dataset is specifically curated to evaluate spatial reasoning in T2I models.
It contains 25,280 text prompts describing spatial relationships (\textit{e.g.,} left, right, above, below) between pairs of objects.
The objects are drawn from 80 categories based on the MS-COCO dataset.
Prompts are generated using predefined templates (\textit{e.g.,} ``A [object A] to the left of a [object B]") to ensure linguistic clarity and consistency.
Spatial relationships are uniformly represented across all object pairs, providing a standardized evaluation framework.
For each prompt, multiple images are generated and assessed using VISOR and related metrics, offering insights into model performance on spatial reasoning tasks.

\subsection*{D.2. T2I-CompBench}
We evaluate spatial relationships and attribute binding through T2I-CompBench Framework~\cite{t2icompbench}, which provides a comprehensive evaluation of T2I synthesis performance.

\paragraph{Spatial Alignment}
Spatial relationships serve as a key sub-category for evaluating T2I synthesis.
The benchmark defines spatial relationships between objects using terms such as left, right, top, bottom, next to, near, and on the side of.
For ``left'', ``right'', ``top,'' and ``bottom'', spatial relationships are evaluated by comparing the relative positions of the centers of bounding boxes for two objects in the generated image.
Specifically, an object $A$ is considered to be on the left of object $B$ if:
$x_1 < x_2, |x_1 - x_2| > |y_1- y_2|, $and $\text{mIoU} < 0.1$, where and $(x_1, y_1)$ and $(x_2, y_2)$ represent the center coordinates of objects $A$ and $B$, respectively.
For ``near to'', ``near'', and ``on the side of,'' these relationships are determined based on the distances between bounding box centers of two objects relative to a predefined threshold.
To detect objects and determine their spatial positions, UniDet~\cite{unidet}, a pre-trained object detection model, is utilized.

\paragraph{Attribute Binding}
Attribute binding in T2I-CompBench evaluates whether attributes such as color, shape, and texture are correctly associated with the corresponding objects in the generated images.

\begin{itemize}
    \item Texture Binding:
    Assesses the model’s ability to associate texture descriptors (\textit{e.g.,} ``fluffy,'' ``metallic'') with the correct objects. 
    Prompts such as ``A rubber ball and a plastic bottle'' test texture-related attribute binding. Texture descriptors are generated from predefined attributes, including ``wooden'', ``glass'', and ``fabric''.
    \item Color Binding:
    Evaluates whether colors are correctly assigned to the objects mentioned in the prompt.
    For example, the prompt ``A blue backpack and a red bench'' tests whether the correct colors are applied to the respective objects.
    Color confusion is a common issue when multiple objects and attributes coexist within a prompt.
    \item Shape Binding:
    Focuses on correctly binding shape descriptors (\textit{e.g.,} ``rectangular'', ``circular'') to objects.
    Prompts such as ``An oval sink and a rectangular mirror'' evaluate shape-related accuracy.
    Shape descriptors include common geometric terms such as ``cubic,'' ``pyramidal,'' and ``circular''.
\end{itemize}
The evaluation utilizes the BLIP-VQA~\cite{li2022blip} model for a fine-grained assessment of object attribute alignment.
BLIP-VQA takes a generated image as input and answers questions about object-attribute pairs (\textit{e.g.,} ``A green bench?", ``A red car?").
The model assigns probabilities to each answer (``Yes'' or ``No''), which are used to compute an overall attribute-binding score.
The final score is calculated as the produce of the probabilities for all attribute-related questions:
$\text{score} = P(\text{`A green bench?'}) \times P(\text{`A red car?'}).$
T2I-CompBench systematically evaluates the model's capability to handle both spatial relationships and attribute binding by providing structured text prompts and analyzing whether the generated images meet the specified constraints.

\paragraph{Dataset}
T2I-CompBench is a benchmark consisting of 6,000 text prompts, generated using predefined templates and ChatGPT~\cite{brown2020language}. 
Each sub-category (\textit{e.g.,} Color, Shape, Texture) includes 1,000 prompts, with 700 used for training and 300 for testing.

\subsection*{D.3. TIFA}
The TIFA (Text-to-Image Faithfulness Evaluation) metric~\cite{hu2023tifa} is designed to measure the alignment between generated images and their corresponding input text prompts.
Unlike traditional metrics such as the FID score, which primarily evaluates the visual quality of images, TIFA emphasizes semantic consistency, assessing whether the content of an image faithfully represents the objects, attributes, and relationships described in the text.
TIFA operates by generating targeted questions based on the input text, leveraging large language models such as LLaMA2~\cite{touvron2023llama} to identify key objects, attributes, and spatial relationships. These questions are designed to ensure alignment between the image and the prompt.
For example, given the text prompt ``A red apple to the left of a green mug,'' the model generates queries such as, ``What color is the apple?'' or ``What object is on the left of the mug?''.
The generated questions are then directed at the output image using a Visual Question Answering (VQA) system.
TIFA typically employs advanced VQA models, such as Owl-ViT~\cite{minderer2022simple} or BLIP (Bootstrapped Language-Image Pretraining)~\cite{li2022blip} to extract objects and attributes from the image and provide answers to the posed questions.
At this stage, the system maps detected objects and their attributes in the image to the corresponding text-based questions, ensuring semantically relevant responses.
Finally, TIFA evaluates the degree of alignment between the generated answers and the expected responses inferred from the input text. High semantic accuracy results in higher scores, while inconsistencies lead to lower scores.
Through this process, TIFA quantitatively measures the semantic fidelity between text and image, offering a robust assessment of how well a model adheres to textual descriptions during image generation.

\paragraph{Dataset.}
The dataset used to compute this metric is based on the same datasets utilized by preceding models~\cite{attend_excite, divide_bind, guo2024initno, conform}.
The text prompts fall into three categories: (1) ``a [animal A] and a [animal B]'', (2) ``a [animal] and a [color][object],'', and (3) ``a  [colorA][objectA] and a [colorB][objectB]''.
These prompts are constructed using 12 animals, 12 objects, and 11 colors.
Each prompt incorporates a subject-color combination, with colors randomly assigned to each subject.
This process results in 66 combinations for animal-animal and object-object pairs, along with 144 animal-object pairs.
Each prompt is then used to generate 64 images with 64 random seeds, ensuring a diverse evaluation of model performance.

\subsection*{D.4. User Studies}
We conducted a user study to evaluate our STORM model based on their ability to generate images that align with detailed text prompts. We created 10 custom prompts, each describing specific objects, attributes, and spatial relationships. 
Using different random seeds, we generated corresponding images with various T2I training-free models. These images were evaluated by 30 participants, who rated them on a scale from 1 (lowest) to 5 (highest) across four criteria; (1) object accuracy, (2) attribute matching, (3) spatial correctness, (4) overall fidelity.
The total score for each model within a given criterion was obtained by summing the scores across all participants. To compare performance, we calculated the percentage score of the \(i\)-th model as the ratio of its total score of all models evaluated within that criterion, using the following formula:

\begin{equation}
\text{Percentage Score (\%)} = \left( \frac{S_i}{\sum_{j=1}^{N} S_j} \right) \times 100
\end{equation}
where \( S_i \) represents the total score of the \(i\)-th model within a given criterion, and \( \sum_{j=1}^{N} S_j \) is the sum of the total scores for all models within that criterion, and \( N \) is the total number of models.
For example, in the spatial correctness criterion, the total scores were 460 for SD, 514 for Attend\&Excite, 507 for Divide\&Bind, 512 for INITNO, s for CONFORM, and 1399 for STORM. The percentage score for STORM in this criterion was calculated as:
\begin{equation}
\frac{1399}{460 + 514 + 507 + 512 + 503 + 1399} \times 100 \approx 35.92\%
\end{equation}
This process was repeated for each criterion and model, with results rounded to the third decimal place.

\section*{E. Additional Experiments}
\subsection*{E.1. Additional Ablation Study}
We construct an additional ablation study on varying $\omega$ values (Section E.1.1) and the effects of applying STO for shorter durations (Section E.1.2).

\subsubsection*{E.1.1. Ablation Studies for $\omega$}
This ablation study examines the impact of using fixed values of $\omega$ compared to the dynamic adjustment employed in our approach.
In our STORM model, $\omega$ dynamically updated across timesteps, as defined in Eq.~\eqref{eq:omega}.
This function enables $\omega$ to gradually increase throughout the diffusion process, maintaining a balance between spatial flexibility in the early steps and precise alignment in later steps.
Table~\ref{table:omega} presents an evaluation of VISOR~\cite{visor} under different values of $\omega$.
The first row, which corresponds to a low $\omega$, demonstrates poor performance in both OA(\%) and VISOR metrics, particularly in $\text{VISOR}_4$.
The moderate $\omega = 50$ and $\omega=100$ show the improvement in OA and VISOR metrics compared to low $\omega$.
However, these fixed values of $\omega$ fail to capture the optimal balance across timesteps, as seen in the lower scores for $\text{VISOR}_4$ when compared to our dynamically adjusted $\omega(t)$.
Specifically, a high $\omega=100$ over-penalizes deviations, reducing the flexibility needed in early timesteps, while moderate $\omega = 50$ does not provide sufficient precision in later timesteps.
In contrast, our dynamically adjusted $\omega(t)$ achieves the best performance across all metrics. 
By gradually increasing $\omega$ throughout the diffusion process, our model effectively balances early-stage flexibility with late-stage spatial accuracy.
This dynamic adjustment leads to superior results, by the significant improvements in $\text{VISOR}_4$ (25.70\%) and $\text{VISOR}_\text{uncond}$ (57.58\%).

\begin{table}[t!]
\centering
\caption{Ablation study on $\omega$ values, comparing fixed settings ($\omega = 1, 50, 100$) with our dynamically adjusted $\omega(t)$, evaluated on OA(\%) and VISOR metrics. It shows lower $\omega$ performs poorly, while moderate and high fixed $\omega$ improve alignment. Our dynamic $\omega(t)$ achieves the best performance by balancing flexibility in early timesteps and precision in later timesteps.}
\label{table:omega}
\resizebox{\columnwidth}{!}{ 
\begin{tabular}{c ccccccc}
\toprule
\multirow{2}{*}{Values of $\omega$} & \multirow{2}{*}{OA (\%)} & \multicolumn{6}{c}{VISOR} \\ \cmidrule(lr){3-8}
    &    & uncond & cond  & 1   & 2  & 3  & 4     \\ \midrule
1  & 39.05  & 33.57 & 85.98 & 66.45 & 41.40 & 20.09 & 6.52  \\ 
50  & 60.73  & 55.65  & 91.64 & 83.81 & 69.09 & 47.82 & 21.91   \\ 
100  & 58.67  & 54.67  & 93.18 & 83.62 & 67.74 & 45.89 & 21.63   \\
\rowcolor{blue!5} 
Ours  & \textbf{61.01}  &  \textbf{57.58} & \textbf{94.39} & \textbf{85.93} & \textbf{69.71} & \textbf{49.01} & \textbf{25.70}    \\
\bottomrule
\end{tabular}
}
\end{table}

\subsubsection*{E.1.2. Ablation Studies for applying STO through Timestep}
In Table~\ref{table:timestep2}, we present an ablation study evaluating the effect of applying STO over different timestep ranges.
In the main paper, STO was applied during the later stages of generation, specifically in the ranges 19–24, 13–24, 7–24, and 1–24, focusing on its impact when image details are refined (see Fig.~\ref{fig:ablation_sup} for more results).
Here, we shift our attention to earlier timesteps, applying STO in the ranges 1-6 (Exp.\#1), 1-12 (Exp.\#2), 1-18 (Exp.\#3), and 1-24 (Exp.\#4). 
This allows us to examine its effectiveness during the early stages of generation, where the model primarily establishes the structural layout and ensures broader spatial consistency.
As shown in Table~\ref{table:timestep2}, optimizing over a longer timestep range yields better results than optimizing over a smaller range.
However, when analyzing the overall VISOR scores, we observe that they are significantly higher than those presented in Table 4 of the main paper. 
This suggests that applying STO during the early timesteps is particularly beneficial, as it enables better spatial adjustments in the initial stages of generation, ultimately leading to improved overall performance.

\begin{table}[t!]
\centering
\caption{Ablation study on the impact of applying STO at different timesteps. Exp.\#A0 represents the baseline results from SD without STO. From Exp.\#A1 to Exp.\#A4, STO is progressively applied over increasing timestep ranges: 1–6, 1–12, 1–18, and 1–24.}
\label{table:timestep2}
\resizebox{\columnwidth}{!}{ 
\begin{tabular}{c ccccccc}
\toprule
\multirow{2}{*}{\#Exp.} & \multirow{2}{*}{OA (\%)} & \multicolumn{6}{c}{VISOR} \\ \cmidrule(lr){3-8}
                  &                          & uncond & cond  & 1     & 2     & 3     & 4                         \\ \midrule
\rowcolor{gray!20} 
0 (SD) & 29.86 & 18.81 & 62.98 & 46.60 & 20.11 & 6.89 & 1.64 \\
1  & 49.00  & 43.45 & 88.67 & 75.92 & 53.53 & 31.70 & 12.71  \\ 
2  & 56.17  & 51.33  & 91.37 & 81.62 & 63.56 & 41.35 & 18.92   \\ 
3  & 59.05  & 54.30  & 91.96 & 82.73 & 66.29 & 45.64 & 22.74   \\
\rowcolor{blue!5}
4 (Ours) & \textbf{61.01}  &  \textbf{57.58} & \textbf{94.39} & \textbf{85.93} & \textbf{69.71} & \textbf{49.01} & \textbf{25.70}    \\
\bottomrule
\end{tabular}
}
\end{table}

\subsection*{E.2. Additional Qualitative Results}
\subsubsection*{E.2.1. Synergy with Stronger Text Encoder}
\begin{figure}[h!]
  \centering
  \includegraphics[width=\linewidth]{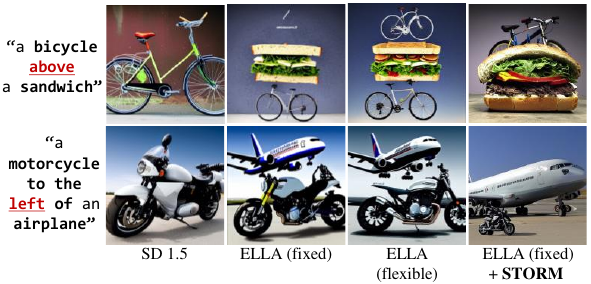}   \caption{Comparison of results from ELLA and ELLA + STORM using SD 1.5.}
  \label{fig:ella}
\end{figure}

Powerful text encoders have been proposed to address spatial alignment in T2I synthesis, with ELLA~\cite{hu2024ella} being a prominent example. 
We leverage STORM’s training-free characteristic to integrate it with ELLA~\cite{hu2024ella}, and we experimentally validate the effectiveness of this combination on the VISOR benchmark. 
As shown in Table~\ref{table:ella}, adding STORM (training-free) to ELLA achieves noticeably better results. 
Fig.~\ref{fig:ella} further provides a visual demonstration of this improvement, illustrating how our training-free approach can complement advanced text-encoder-based methods for spatial alignment.

\begin{table}[!h]
\small
\caption{Quantitative results on the VISOR benchmark by combining STORM with ELLA.}
\label{table:ella}
\resizebox{\columnwidth}{!}{ 
\begin{tabular}{lccc}
\toprule
model &OA & VISOR (cond) & VISOR (uncond) \\ \hline
SD 1.5 & 28.49&62.94&17.93 \\
SD 1.5 + ELLA (fixed)& 52.7 & 67.31 & 35.48 \\
SD 1.5 + ELLA (flexible) & 54.33 & 67.51 & 36.68 \\
\rowcolor{blue!5}
SD 1.5 + \textbf{STORM} & \textbf{62.03} & \textbf{90.82} & \textbf{56.33} \\
\bottomrule
\end{tabular}}
\end{table}

\subsection*{E.3. Additional Visualization}
We provide additional visualizations to further support our results. These include comparisons between Stable Diffusion (SD) and our method on the SR2D dataset (Section E.2.1), qualitative results for both SD 1.4 and SD 2.1 (Section E.2.2), visualizations of attention maps across denoising timesteps (Section E.2.3), additional ablation visualizations illustrating the effect of applying STO during the denoising process (Section E.2.4), and positional variations observed within the same seed (Section E.2.5).

\textit{Note.} Full-page figures are placed at the bottom of the document.

\subsubsection*{E.3.1. Comparison between SD and Ours} 
As illustrated in Fig.~\ref{fig:fig_sd}, we provide additional visualization on stable diffusion~\cite{sd} and ours using SR2D Dataset~\cite{visor}.
Our model, STORM, demonstrates a remarkable ability to accurately position objects in the desired locations.

\subsubsection*{E.3.2. Additional Qualitative Results}
In Fig.~\ref{fig:sd1_4_1} and Fig.~\ref{fig:sd1_4_2}, we present the qualitative comparisons between Stable Diffusion 1.4~\cite{sd} and other state-of-the-art methods~\cite{conform, guo2024initno, divide_bind, attend_excite}.
Our model excels in accurately matching attributes while ensuring that all objects are distinctly generated without overlaps.
Moreover, unlike other methods that often struggle with positional accuracy, our approach consistently maintains precise spatial arrangements, demonstrating superior performance.
To further validate our findings, we conducted the same qualitative comparisons across all methods using Stable Diffusion 2.1~\cite{sd}. As shown in Fig.~\ref{fig:sd2_1_1} and Fig.~\ref{fig:sd2_1_2}, our method continues to exhibit strong spatial understanding, regardless of the model version.
Additionally, it effectively mitigates object overlap issues, a common weakness in Stable Diffusion, further highlighting its robustness in generating well-structured outputs.

\subsubsection*{E.3.3. Additional Visualization for attention map}
We provide extended visualization of attention map progression throughout the denoising process in Fig.~\ref{fig:attn1} and Fig.~\ref{fig:attn2}.
The figure on the far left shows the attention map at the initial stages of the denoising process, while the subsequent figures to the right illustrate the attention maps as the denoising progresses through later steps.
As shown in the figures, although both models start with the same noise distribution, they gradually exhibit different patterns.
Notably, our model exhibits a clear tendency to focus on regions requiring refinement, ensuring a precise distribution in those areas. 
In contrast, Stable Diffusion often displays scattered attention distributions, with some cases showing complete dissipation of attention in certain regions.

\subsubsection*{E.3.4. Applying STO During the Denoising Process}
\begin{figure*}[h!]
  \centering
  \includegraphics[width=\textwidth]{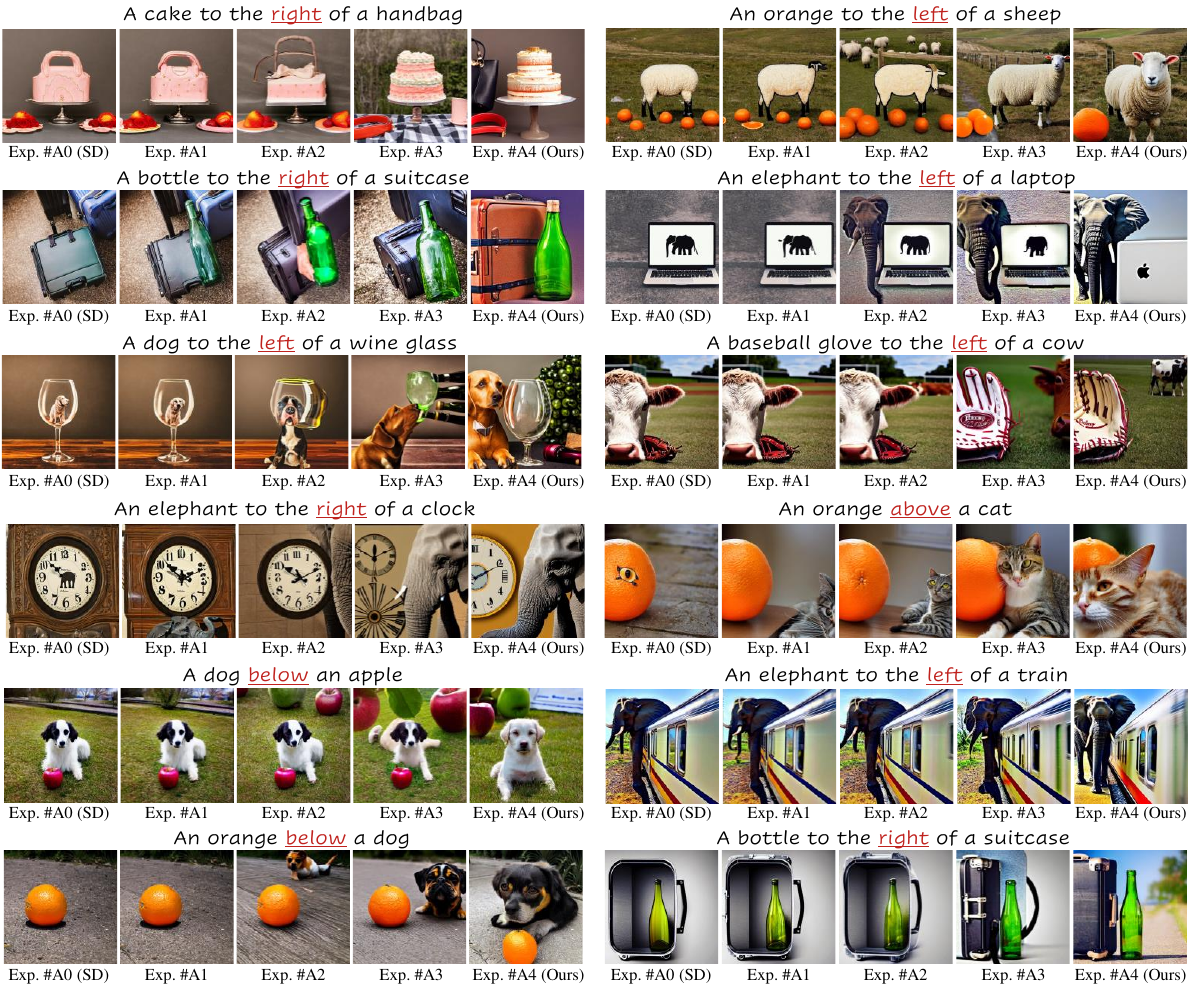}
  \caption{Additional comparison of results when applying STO at different timesteps. Experiments are organized as follows: no STO (Exp.\#A0), STO applied from timesteps 19-24 (Exp.\#A1), 13-24 (Exp.\#A2), 7-24 (Exp.\#A3), and 1-24 (Exp.\#A4). As seen in the images, earlier STO application improves object positioning and reduce overlap, resulting in more accurately positioned objects.} 
  \label{fig:ablation_sup}
\end{figure*}

In Fig.~\ref{fig:ablation_sup}, we provide additional visualizations for the ablation study, demonstrating the impact of applying STO during the denoising process.

\subsubsection*{E.3.5. Additional Visualization of Positional Variations}
\begin{figure*}[h!]
  \centering
  \includegraphics[width=\textwidth]{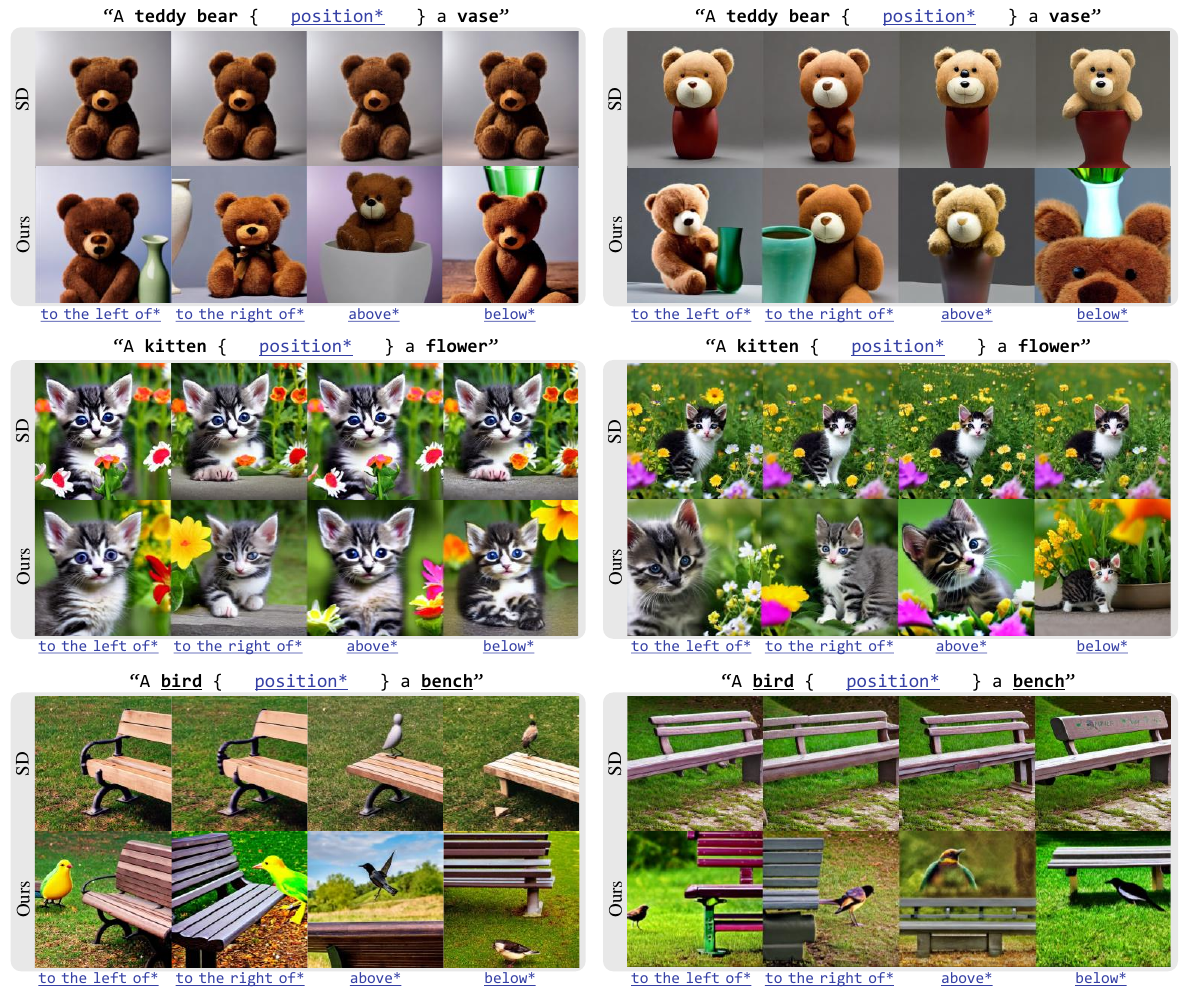}
  \caption{Additional Comparison of Spatial Awareness. \{ position* \} in each prompt denotes the spatial relationship in each column (\textit{e.g.,} ``to the left of'', ``to the right of'', ``above,'' and ``below'').}
  \label{fig:seed}
\end{figure*}

To further validate the effectiveness of our method in understanding and reflecting spatial prompts, we provide additional examples demonstrating the model's spatial awareness across various object combinations within the spatial relationships.
As shown in Fig.~\ref{fig:seed}, SD exhibits limited spatial awareness, often generating nearly identical images regardless of the given spatial prompts.
In contrast, our model effectively captures and preserves the specified spatial relationships, demonstrating a superior understanding of spatial constraints.

\subsection*{E.4. Experiments on Complex, Diverse, and 3D Positional Prompts}
\begin{figure*}[h!]
  \centering
  \includegraphics[width=\textwidth]{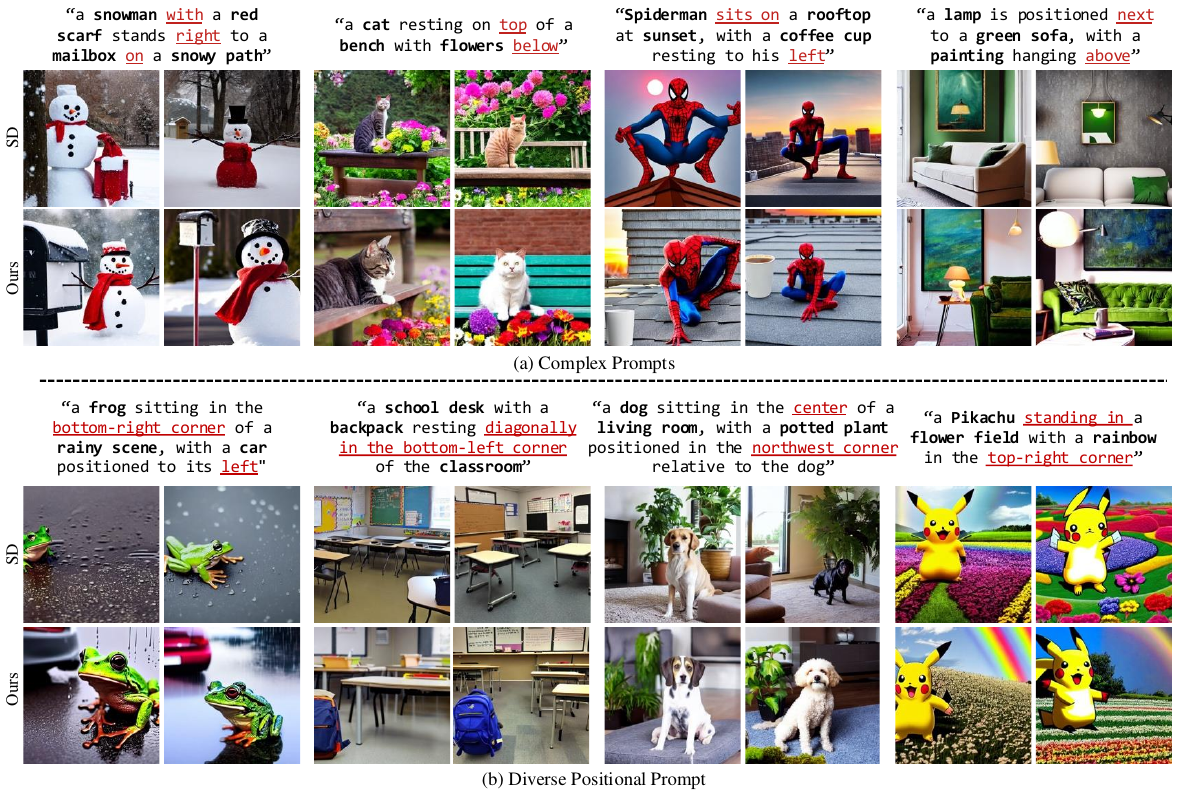}
  \caption{Additional results on (a) Complex Prompts: prompts with three or more spatial relationships, and (b) Diverse Positional Prompts: including not only left, right, above, and below but also diagonal spatial relationships. Our model successfully captures these complex and diverse spatial constraints, outperforming Stable Diffusion.}
  \label{fig:complex}
\end{figure*}

To further evaluate the robustness and versatility of our model, we conducted additional experiments using (a) complex prompts and (b) diverse positional prompts, as shown in Fig.~\ref{fig:complex}.
Complex prompts involve three or more spatial relationships, requiring the model to accurately interpret and generate objects while maintaining overlapping or hierarchical spatial constraints.
These experiments demonstrate that our method effectively captures spatial alignment across a wide range of challenging and diverse scenarios.
Additional results include (a) Complex Prompts, where prompts contain three or more spatial relationships, and (b) Diverse Positional Prompts, which extend beyond left, right, above, and below to include diagonal spatial relationships. Our model successfully captures these intricate spatial constraints, consistently outperforming Stable Diffusion.
We also generate images using 3D positional prompts, as shown in Fig.~\ref{fig:3d}. 
Although our approach is fundamentally designed for 2D spatial reasoning, resulting in slightly fewer natural outcomes compared to 2D scenarios, it significantly outperforms other models that entirely disregard positional cues, demonstrating significantly better generation quality.

\section*{F. Discussion}
\subsection*{F.1. Failure Cases}
\begin{figure*}[h!]
  \centering
  \includegraphics[width=0.8\textwidth]{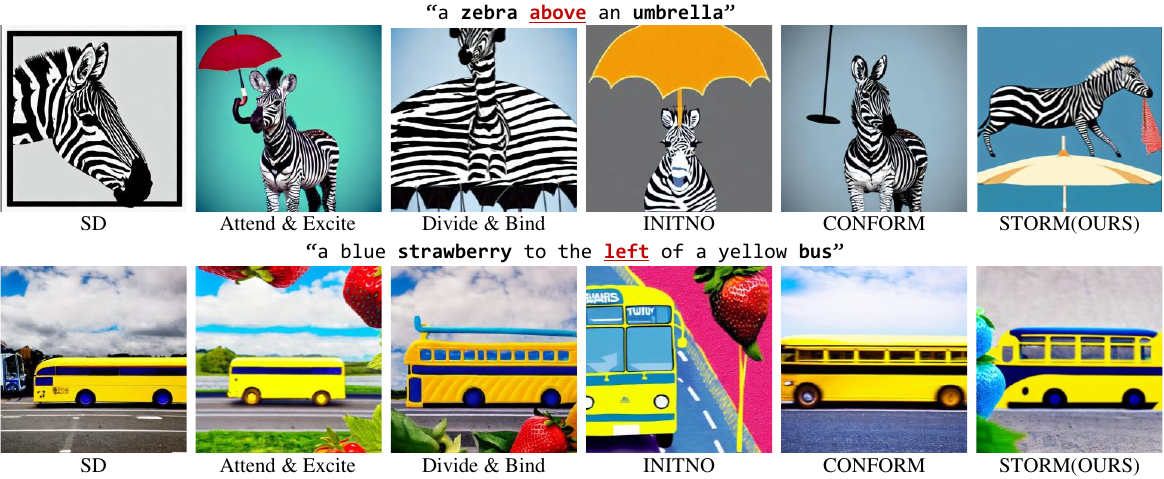}
  \caption{Limitations. Difficulty handling rare combinations of objects and attributes.} 
  \label{fig:limitation}
\end{figure*}
Despite achieving remarkable performance in spatial alignment, our method faces challenges with extremely rare object-attribute combinations and positional prompts requiring three-dimensional spatial reasoning.
These difficulties arise from the training-free nature of our approach, which inherently suffers from data biases and lacks exposure to such uncommon scenarios.
For instance, as shown in the second row of Fig.~\ref{fig:limitation}, the model effectively generates common objects like a ``yellow bus''.
However, it struggles with rarer combinations such as a ``blue strawberry,'' resulting in either failed generations or outputs with significantly lower image quality.
While our method is capable of producing a ``blue strawberry,'' it exhibits a noticeable degradation in overall image fidelity.
Similarly, as seen in the first row of Fig.~\ref{fig:limitation}, placing a ``zebra'' on the top of an ``umbrella'' leads to an unnatural and awkward composition, highlighting the difficulty of generating plausible outputs for spatially improbable scenarios.
Furthermore, our model is currently designed to reason within a 2D space, effectively capturing relationships such as left, right, above, and below.
However, since 3D positional cues are not explicitly considered, the generations for 3D prompts can sometimes appear less natural or accurate compared to their 2D counterparts (see Fig.~\ref{fig:3d}).
Despite these limitations, our method consistently outperforms others that do not account for spatial relationships, delivering superior results overall.

\subsection*{F.2. Future Works}
Our proposed STORM framework effectively mitigates spatial misalignment in training-free T2I synthesis, paving the way for several promising research directions.
One key avenue for future work is extending STORM to support multimodal inputs, such as integrating audio or video cues within text-based prompts. 
This would enhance the model’s adaptability across diverse creative applications. 
Additionally, optimizing the computational efficiency of STO could enable real-time applications, including interactive art and game design.
Another promising direction involves developing methods for dynamically optimizing image generation based on immediate user input, allowing for greater flexibility and responsiveness.
While this study primarily focuses on relative positioning (e.g., left, right, above, below), future research could explore more complex spatial relationships, such as 3D spatial reasoning and multi-object interactions.
Although our model already demonstrates strong performance in these areas, we believe there is substantial potential for further advancements that could push the boundaries of spatially aware text-to-image generation.

\begin{figure*}[h!]
  \centering
  \includegraphics[height=\textheight]{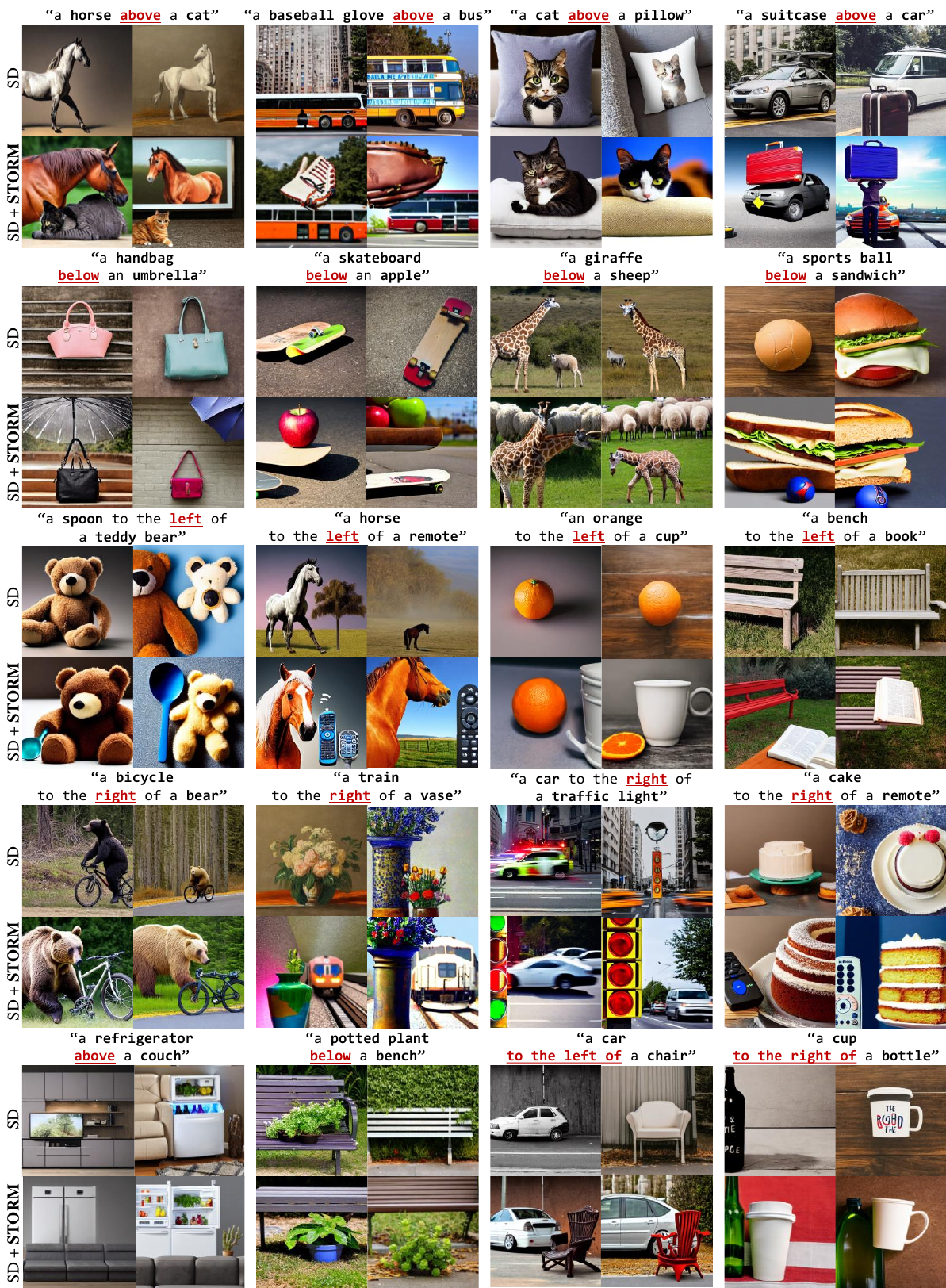}
  \caption{Comparison between Stable Diffusion 1.5 and ours on SR2D Dataset~\cite{visor}.} 
  \label{fig:fig_sd}
\end{figure*}
\clearpage

\begin{figure*}[h!]
  \centering
  \includegraphics[width=\textwidth]{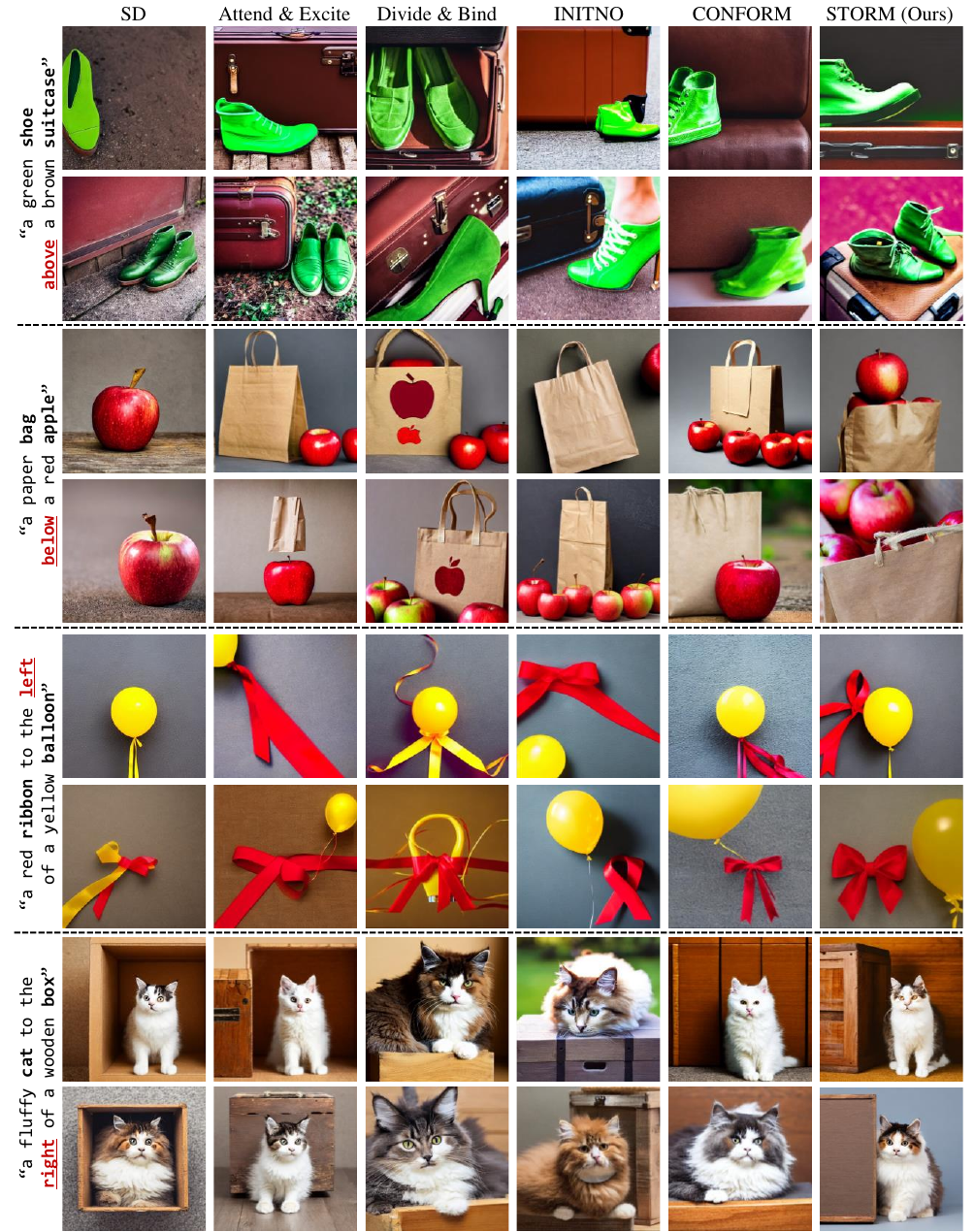}
  \caption{Additional qualitative results using \textit{Stable Diffusion 1.4} in comparison with state-of-the-art methods.} 
  \label{fig:sd1_4_1}
\end{figure*}
\clearpage

\begin{figure*}[h!]
  \centering
  \includegraphics[width=\textwidth]{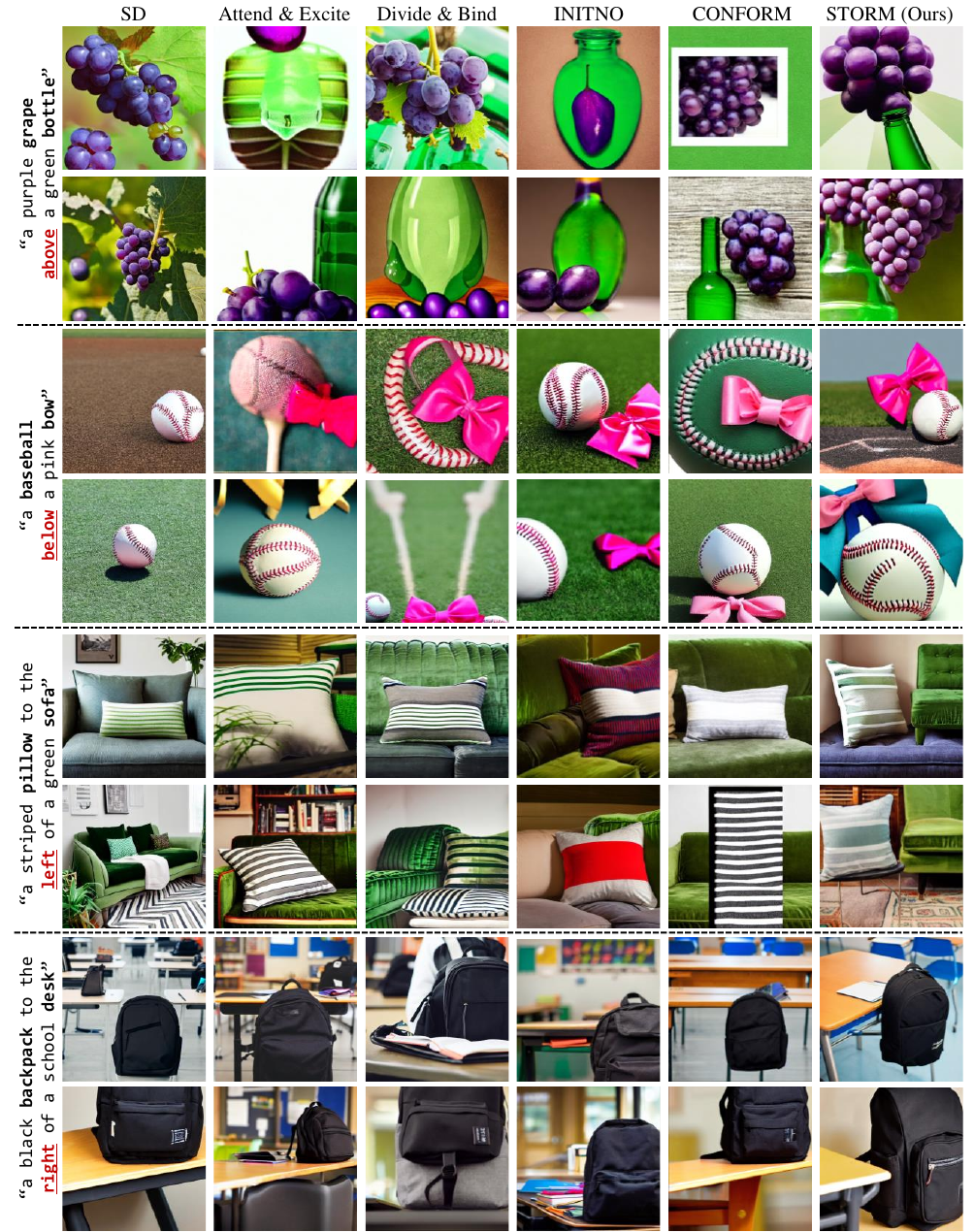}
  \caption{Additional qualitative results using \textit{Stable Diffusion 1.4} in comparison with state-of-the-art methods.} 
  \label{fig:sd1_4_2}
\end{figure*}
\clearpage

\begin{figure*}[h!]
  \centering
  \includegraphics[width=\textwidth]{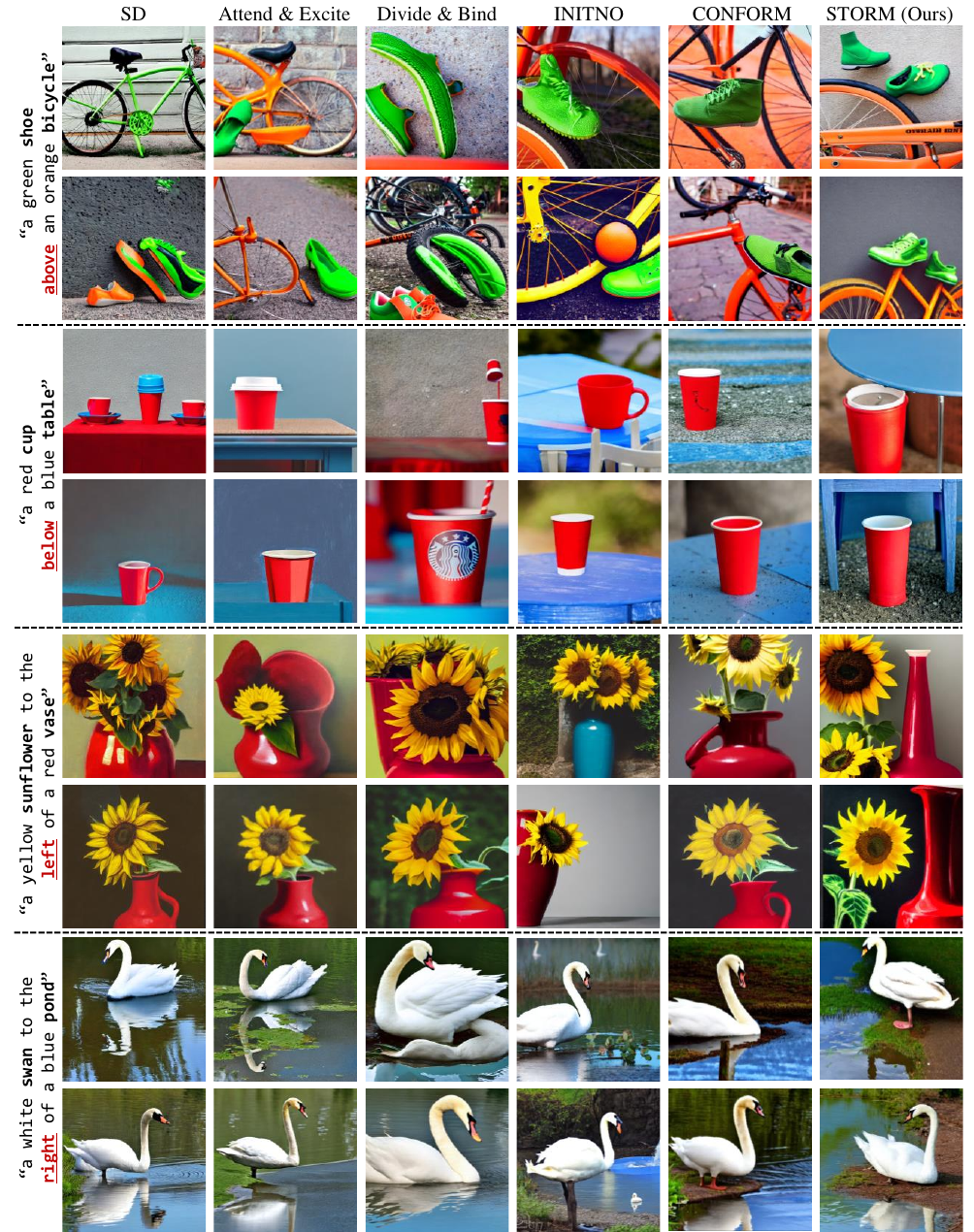}
  \caption{Additional qualitative results using \textit{Stable Diffusion 2.1} in comparison with state-of-the-art methods.} 
  \label{fig:sd2_1_1}
\end{figure*}
\clearpage

\begin{figure*}[h!]
  \centering
  \includegraphics[width=\textwidth]{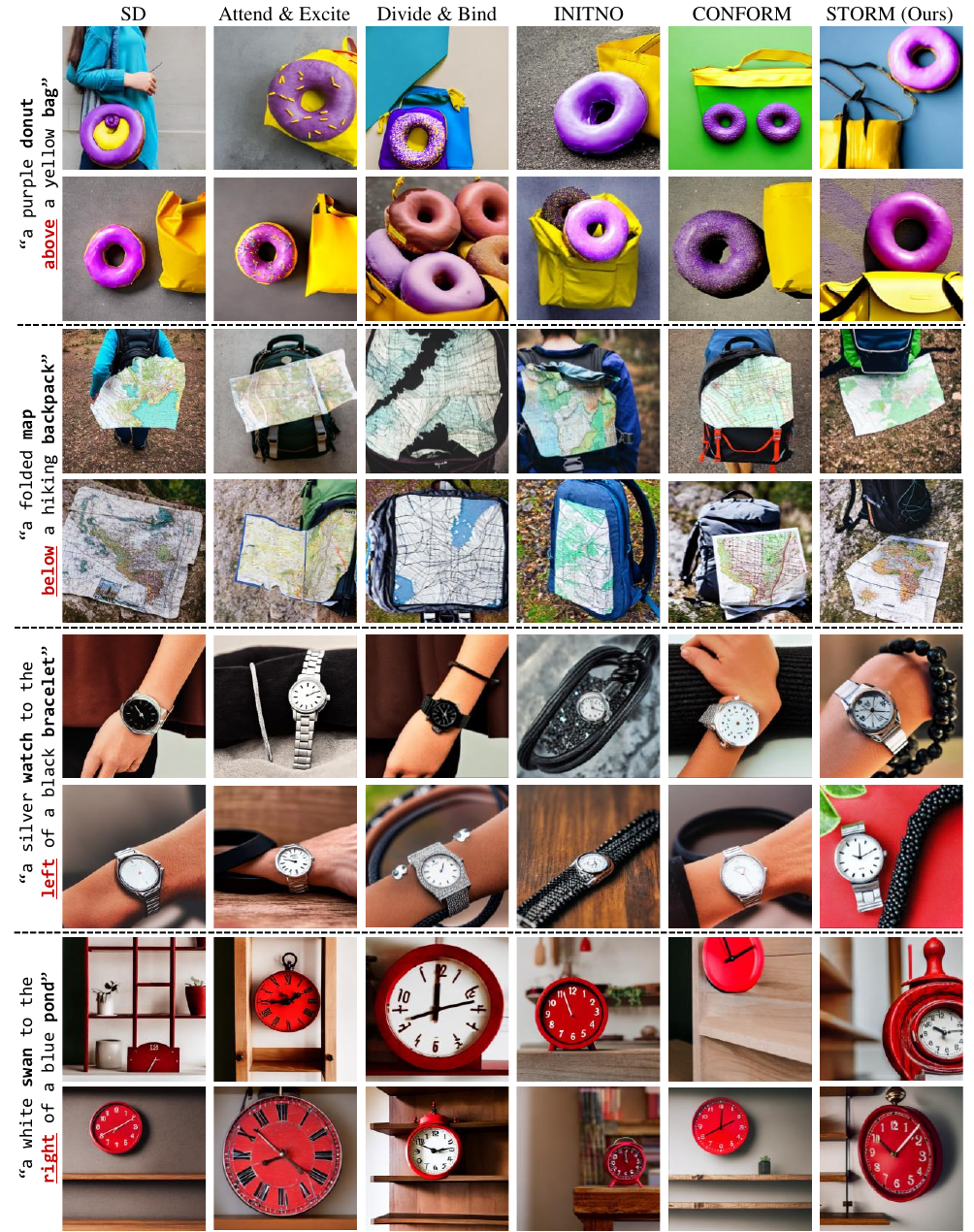}
  \caption{Additional qualitative results using \textit{Stable Diffusion 2.1} in comparison with state-of-the-art methods.} 
  \label{fig:sd2_1_2}
\end{figure*}
\clearpage

\begin{figure*}[h!]
  \centering
  \includegraphics[width=\textwidth]{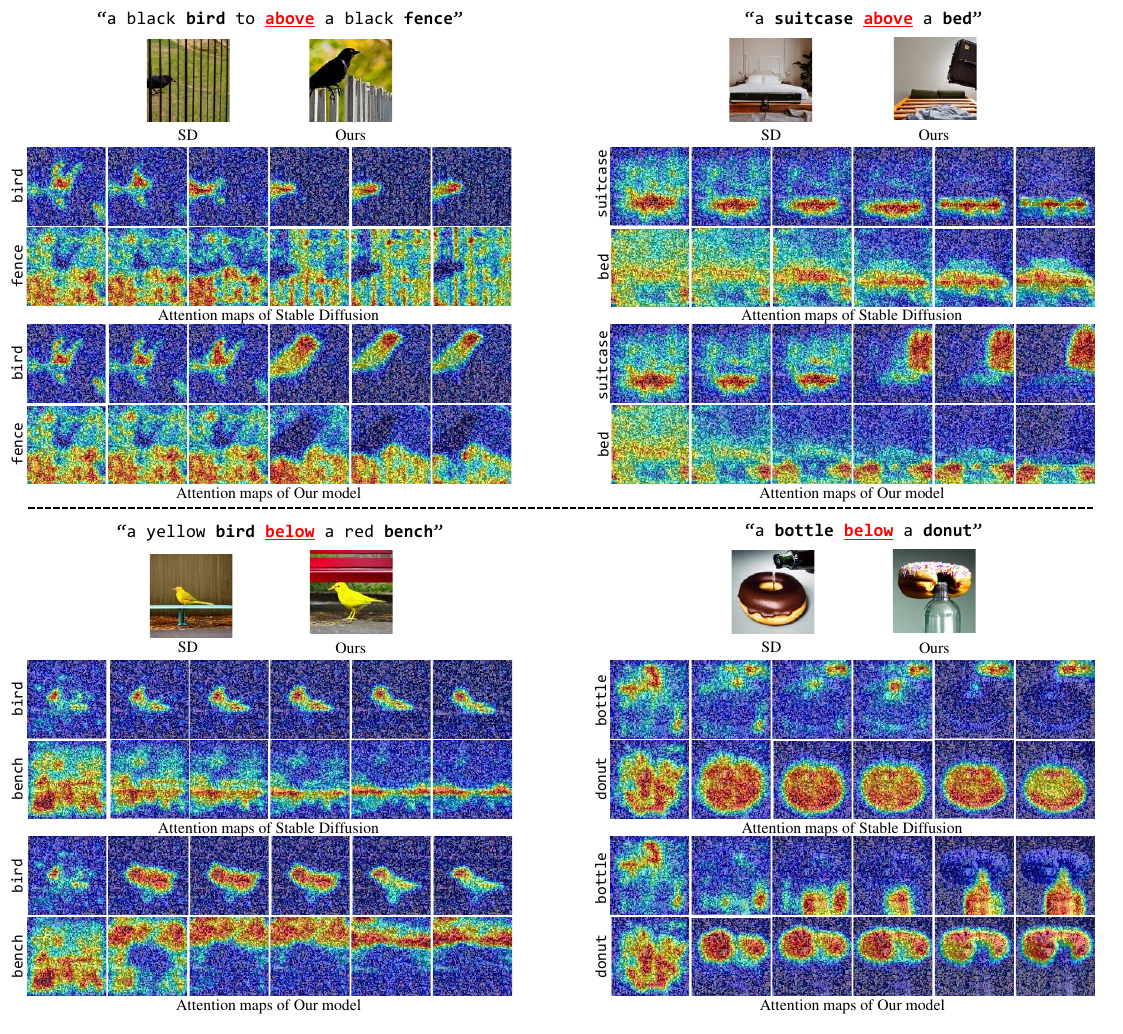}
  \caption{Additional visualizations of the attention map across different denoising timesteps. The leftmost figure represents the visualization of the attention map at the very early denoising steps, and as we move to the right, the figures show the attention maps after progressively more denoising steps.}
  \label{fig:attn1}
\end{figure*}
\clearpage

\begin{figure*}[h!]
  \centering
  \includegraphics[width=\textwidth]{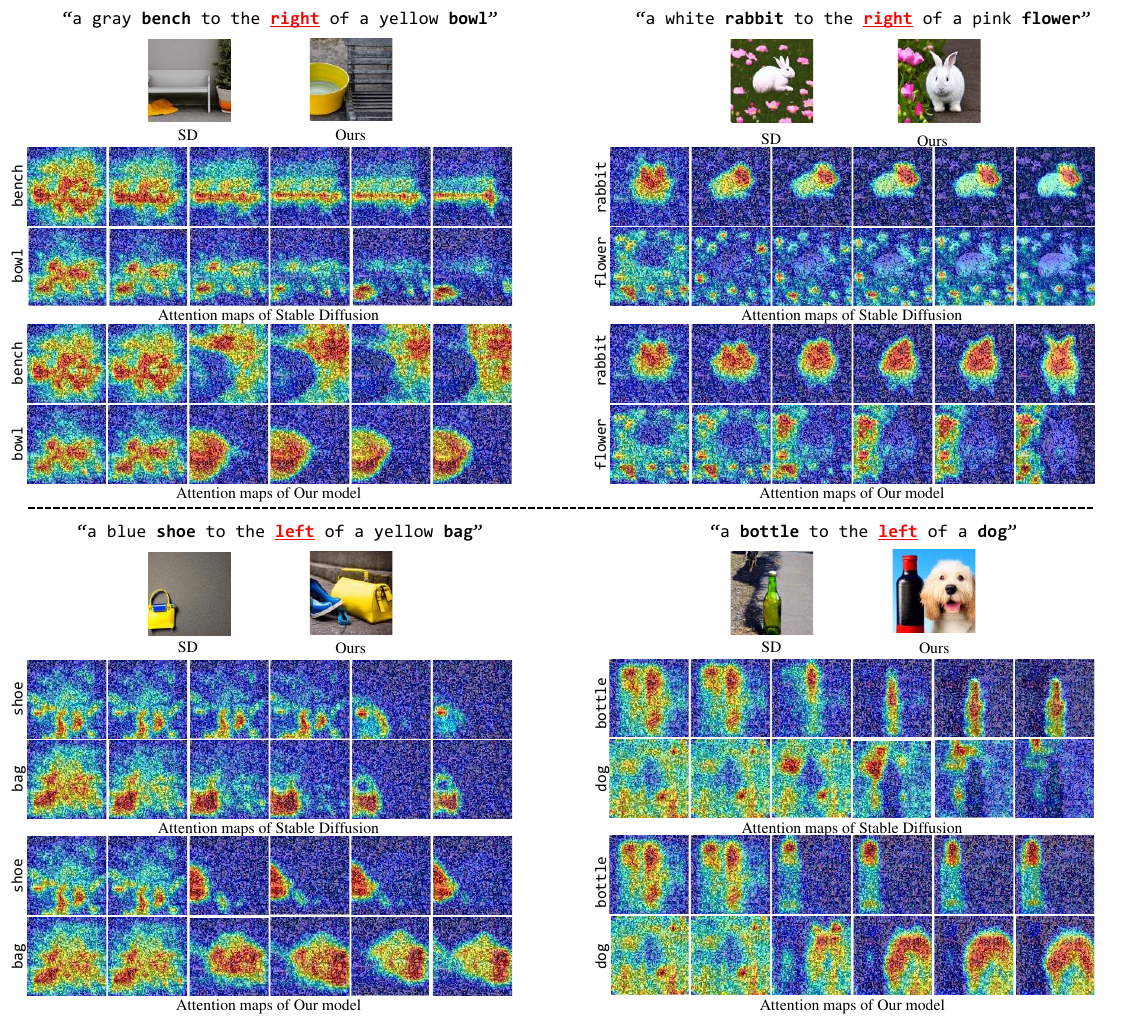}
  \caption{Additional visualizations of the attention map across different denoising timesteps. The leftmost figure represents the visualization of the attention map at the very early denoising steps, and as we move to the right, the figures show the attention maps after progressively more denoising steps.}
  \label{fig:attn2}
\end{figure*}
\clearpage

\begin{figure*}[h!]
  \centering
  \includegraphics[height=\textheight]{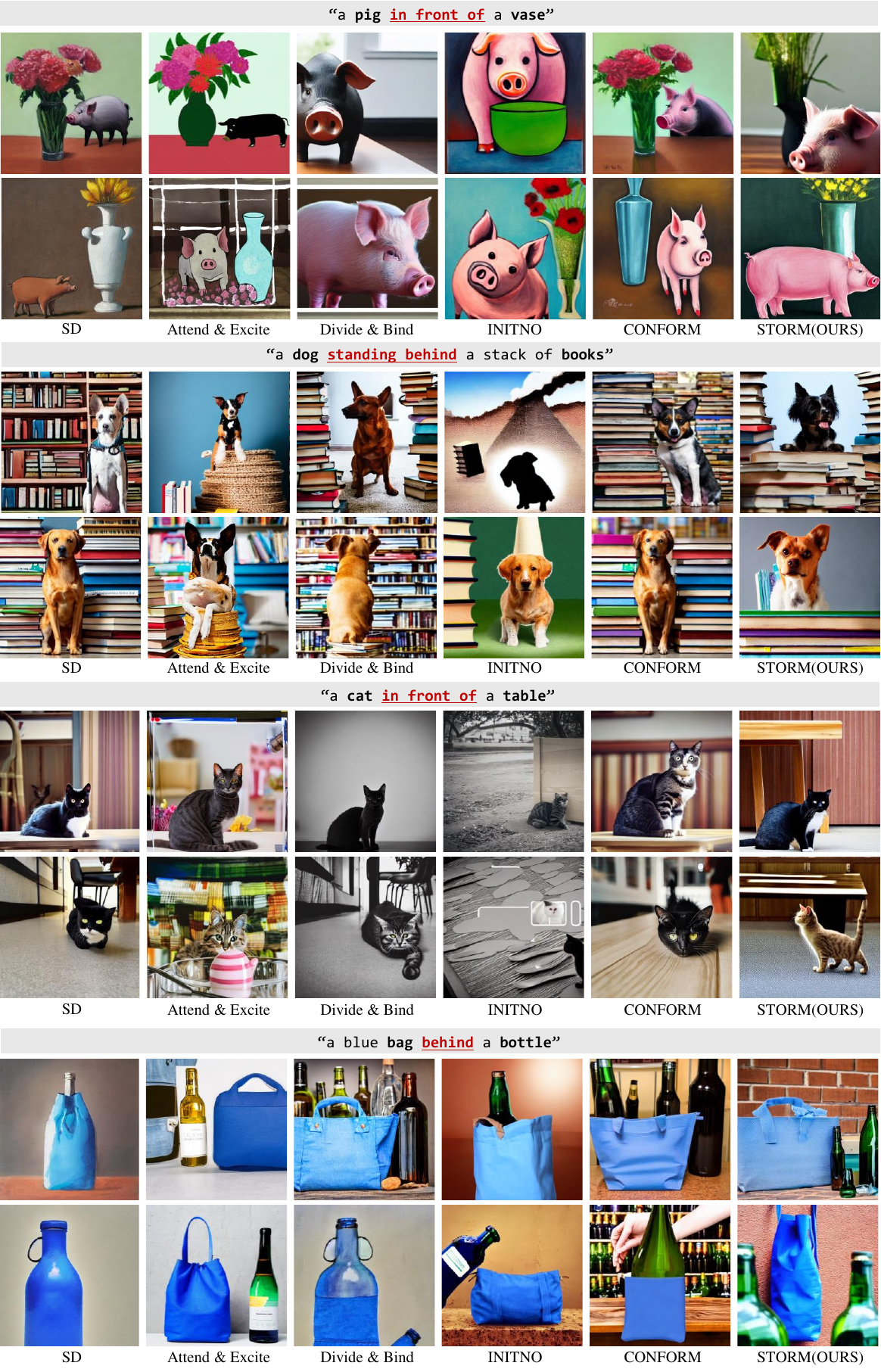}
  \caption{Additional results on 3D positional prompts.}
  \label{fig:3d}
\end{figure*}
\clearpage

\end{document}